\documentclass[10pt,twocolumn,letterpaper]{article}

\usepackage{cvpr}             
\usepackage[accsupp]{axessibility} 

\usepackage{microtype}

\renewcommand{\paragraph}[1]{%
  \vspace{0.5em}%
  \noindent\textbf{#1}\quad
}

\usepackage{soul}
\setuldepth{foobar}

\usepackage{amsmath,amsfonts,bm}

\def\eqref#1{equation~\ref{#1}}

\def\1{\bm{1}}

\DeclareMathAlphabet{\mathsfit}{\encodingdefault}{\sfdefault}{m}{sl}
\SetMathAlphabet{\mathsfit}{bold}{\encodingdefault}{\sfdefault}{bx}{n}

\def\gD{{\mathcal{D}}}

\def\gL{{\mathcal{L}}}

\def\gP{{\mathcal{P}}}

\usepackage[table]{xcolor}
\definecolor{cvprblue}{rgb}{0.21,0.49,0.74}
\usepackage[pagebackref,breaklinks,colorlinks,allcolors=cvprblue]{hyperref}

\def\paperID{42260} 
\def\confName{CVPR}
\def\confYear{2026}

\title{Cloning Deterministic Worlds:\\The Critical Role of Latent Geometry in Long-Horizon World Models}
\vspace{-3mm}
\author{
Zaishuo Xia$^{1}$, Yukuan Lu$^{1}$, Xinyi Li$^{1}$,
Yifan Xu$^{2}$, 
Yubei Chen$^{1,2\dagger}$\\
\normalsize
$^1$UC Davis \quad
$^2$Open Path AI Foundation\\
}

\begin{document}
\maketitle
\begingroup
\renewcommand\thefootnote{}\footnotetext{$^\dagger$Corresponding author. Project page: \url{https://xiafire.github.io/grwm-project-page}}
\endgroup
\begin{abstract}
A world model is an internal model that simulates how the world evolves. Given past observations and actions, it predicts the future physical state of both the embodied agent and its environment. Accurate world models are essential for enabling agents to think, plan, and reason effectively in complex, dynamic settings. However, existing world models often focus on random generation of open worlds, but neglect the need for high-fidelity modeling of deterministic scenarios (such as fixed-map mazes and static space robot navigation). In this work, we take a step toward building a truly accurate world model by addressing a fundamental yet open problem: constructing a model that can fully clone a deterministic world. 1) Through diagnostic experiment, we quantitatively demonstrate that high-fidelity cloning is feasible and the primary bottleneck for long-horizon fidelity is the geometric structure of the latent representation, not the dynamics model itself. 2) Building on this insight, we show that applying temporal contrastive learning principle as a geometric regularization can effectively curate a latent space that better reflects the underlying physical state manifold, demonstrating that contrastive constraints can serve as a powerful inductive bias for stable world modeling; we call this approach Geometrically-Regularized World Models (GRWM). At its core is a lightweight geometric regularization module that can be seamlessly integrated into standard autoencoders, reshaping their latent space to provide a stable foundation for effective dynamics modeling. By focusing on representation quality, GRWM offers a simple yet powerful pipeline for improving world model fidelity.
\end{abstract}

\section{Introduction}

Creating a high-fidelity, interactive clone of an environment purely from observational data has long been a central ambition in artificial intelligence. Such a clone can serve as a simulator for reinforcement learning agents~\citep{hafner2020dreamer,hao2025neural,sutton}, enable task planning in robotics~\citep{mendonca2023structured,unitree}, and facilitate controllable content generation in games~\citep{alonso2024diffusionworldmodelingvisual,oasis2024,bruce2024genie}. World models are the primary tools for achieving this goal: they aim to capture an environment’s dynamics, predict its future states, and simulate its evolution~\citep{dyna,ha2018world,schrittwieser2020mastering,lecun2022path}.

Most current world models focus on open-world settings~\citep{oasis2024,he2025matrix,genie3,sora}, generating unconstrained and often random environments in which each simulation produces a different world. This approach is intended to enhance generalization by exposing agents to diverse training scenarios~\citep{jang2025dreamgen,alonso2024diffusionworldmodelingvisual,hafner2020dreamer,hafner2019learning,bruce2024genie}. However, the emphasis on randomness can lead to unstable dynamics, making such models ill-suited for applications that demand reliable prediction and precise planning in fixed tasks. In these cases, they often produce futures that are merely plausible, not faithful.

Achieving precise world simulation with world models requires overcoming two tightly intertwined challenges:
\begin{itemize}
    \item Representation learning. Exteroceptive sensory data, such as images, are high-dimensional and encode complex, nonlinear mappings from underlying physical processes~\cite{battaglia2018relational,wang2020understanding,tschannen2018recent, arats2021improving, kang2024far}. This makes accurate future-state prediction difficult even under full observability~\cite{ha2018world,hafner2019learning}. For example, a simple spatial translation in the physical world can correspond to a highly nonlinear trajectory in pixel space~\cite{tancik2020fourier,nonlinear1,nonlinear2}. Addressing this requires a representation space that faithfully encodes the underlying physical states while minimizing information loss and noise, thereby simplifying the subsequent dynamics modeling task~\cite{wang2022denoised,fang2024towards,seo2023masked,karlsson2023predictive,battaglia2018relational,zhou2024dinowmworldmodelspretrained}.
    
    \item Dynamics modeling. Even with an optimal representation, the dynamics model must capture a broad range of transition patterns — including 3D transformations, logical rules, causal dependencies, and temporal memory~\cite{dynamics1,dynamics2,dynamics3,dynamics4,hao2025neural,chen2024diffusion,huang2025selfforcing}. The challenge lies in building a unified mechanism that can accurately model all these regularities~\cite{dynamics1,dynamics4,dynamics5,dynamics6}.
\end{itemize}
These challenges are inherently coupled: a poor representation forces the dynamics model to operate in a noisy, entangled latent space, increasing prediction complexity and reducing generalization; conversely, a well-structured representation is of limited utility if the dynamics model cannot capture the full spectrum of transition patterns~\cite{representation1,representation2,tschannen2018recent,blau2019rethinking,kouzelis2025eq,stab}. Progress toward generic neural world simulation therefore demands a co-design of representation learning and dynamics modeling, ensuring that the latent space is both physically meaningful and optimally aligned with the predictive capabilities of the dynamics model~\cite{ha2018world,seo2023masked,hafner2020dreamer,dynamics5,chen2024deep}.

In this work, we focus on the faithful cloning of deterministic environments — settings governed by fixed rules, such as a 3D maze with a static map. In such cases, the objective is not to produce a plausible world, but to reproduce the unique, true trajectory of the environment. We target these environments for three reasons: (1) Their consistent dynamics make them amenable to precise predictive modeling. (2) The absence of stochasticity eliminates uncertainty, enabling rigorous evaluation of a world model’s fidelity. (3) Many important real-world applications — including robot navigation in fixed spaces and game AI operating on static maps — inherently involve deterministic environments. More importantly, our primary goal is not merely to solve the cloning task itself, but to use deterministic cloning as a controlled setting for understanding the fundamental factors that limit long-horizon fidelity in world models.

Yet, even in this controlled setting, we find that achieving accurate long-horizon cloning of simple deterministic environments remains an open challenge. Across all state-of-the-art baselines we evaluated, none were able to maintain fidelity over extended horizons: small prediction errors accumulate rapidly, causing trajectories to diverge from reality after only a few steps. In contrast, when the dynamics model is provided with the environment’s underlying physical states\footnote{We refer to this version as the \emph{oracle model}, since it has access to the true underlying states of the environment.} — rather than high-dimensional exteroceptive signals such as images — it can produce remarkably accurate long-horizon predictions. This observation suggests that the effectiveness of a world model is fundamentally constrained by the structure of its latent representation space, and thus raises a natural question: How can we build a self-supervised representation that aligns with the underlying physical states, enabling stable and accurate long-horizon predictions?

A natural next step is to investigate how to bridge the gap to oracle-level performance through unsupervised representation learning. We show that applying the temporal contrastive learning principle~\cite{foldiak1991learning, wiskott2002slow, cltt3,cltt5,cltt7,klindt2020towards} as a geometric regularization can effectively curate a latent space that better reflects the underlying physical state manifold, demonstrating that contrastive constraints can serve as a powerful inductive bias for stable world modeling; we call this approach Geometrically-Regularized World Models (GRWM). At its core is a lightweight geometric regularization module that can be seamlessly integrated into standard autoencoders, reshaping their latent space to better align with the underlying physical states and provide a stable foundation for effective dynamics modeling. Through extensive experiments, we demonstrate that improved representation geometry directly translates into higher long-horizon fidelity across multiple world models, validating the ``representation matters" hypothesis. GRWM acts as a plug-and-play component that systematically unlocks long-horizon fidelity for state-of-the-art world models. 

In summary, our main contributions are as follows:
\vspace{-1mm}

\begin{enumerate}
\item Through diagnostic experiment, we quantitatively demonstrate that the primary bottleneck for long-horizon fidelity is the geometric structure of the latent representation, not the dynamics model itself. Our ``oracle" model, which achieves near-perfect accuracy using ground-truth states, confirms that high-fidelity cloning is feasible and highlights that discovering these underlying states is a critical path toward perfect world modeling.

\item We demonstrate the use of temporal contrastive principles as a geometric regularizer for a world model’s autoencoder. Our approach, termed Geometrically-Regularized World Models (GRWM), leverages these principles to explicitly constrain the latent space to align with the environment’s underlying state manifold, showing that contrastive constraints can serve as a strong inductive bias for stable and faithful long-horizon world modeling.

\item We formalize the problem of high-fidelity cloning in deterministic environments, shifting the research focus from plausible open-world generation to reproducible fidelity. We also curate dedicated datasets for this task, providing a clearer benchmark for studying world model fidelity.

\end{enumerate}

\section{Related Work}
\paragraph{Video World Models.} Video world models aim to predict future visual observations of an environment conditioned on agent actions, often using latent generative modeling as a central approach~\cite{oasis2024,he2025matrix, parkerholder2024genie2}. Latent generative world models~\citep{seo2023masked,high,ha2018world} first encode observations into a latent space via an autoencoder and then model transitions in this space to predict future frames. This paradigm is widely applicable~\cite{oasis2024,parkerholder2024genie2,genie3,bruce2024genie,sora} and has inspired recent works exploring various enhancements, including interactive controllability~\cite{i1,oasis2024,a2}, architectural innovations for causality~\cite{yin2025causvid,chen2024diffusion} and persistence~\cite{framepack,a1,far,framepackv1,song2025history}, and improved physical accuracy~\cite{p1,p2,p3}. Despite these advances, the predominant focus has been on improving dynamics modeling, e.g., designing more expressive latent transition models~\cite{t1,huang2025selfforcing,song2025history,a1} and adding causal and relational inductive biases~\citep{mendonca2023structured,hao2025neural,chen2024diffusion}. The autoencoder's role is largely engineering: to provide a compressive latent embedding for the dynamics model\footnote{Commonly, the autoencoder is trained with a small KL weight and additional perceptual or adversarial losses~\cite{high}.}, with most work focusing on improving reconstruction quality rather than shaping the latent geometry for stable long-horizon prediction.

Moreover, most video world models target open-world or stochastic environments, emphasizing diversity and plausibility over faithful reproduction of a unique trajectory~\citep{oasis2024,he2025matrix,genie3,bruce2024genie,sora}. In contrast, deterministic environments, where the goal is to precisely clone the true sequence of states, have received less attention. Our work highlights that in such settings, representation quality can be the primary bottleneck: even strong dynamics models fail if the latent space is poorly structured, motivating methods like GRWM that explicitly regularize latent geometry.

\paragraph{Contrastive Learning.} 
Pure reconstruction objectives often yield degenerate representations that ignore temporal cues. Contrastive learning principles~\citep{drlim,oord2018representation,chen2018sparse,chen2020simple,he2020momentum,wang2020understanding,yeh2022decoupled,garrido2022duality,wang2024pose} suggest a remedy: enforce similarity for related samples and repulsion for unrelated ones. Several methods extend this to temporal data, such as Contrastive Learning Through Time (CLTT)~\citep{cltt1,cltt2,cltt3,cltt4,cltt5,cltt6,cltt7,cltt8,wang2024pose}, where contrastive objectives are used to capture temporal coherence in sequential observations.

GRWM builds directly on this foundation. While these methods successfully learn temporally coherent representations, they typically use contrastive objectives directly for representation learning. In contrast, we employ temporal contrastive learning principles as a geometric regularizer within a world modeling framework. This regularization explicitly reshapes the latent space of a world model’s autoencoder to align with the underlying physical state manifold. Through this formulation, we reveal that contrastive constraints can serve as a powerful inductive bias for improving world model fidelity, validating that representation quality is a central factor in achieving stable and accurate long-horizon prediction.
\section{Preliminary}
\textbf{The Next-State Prediction Problem.}  
World models aim to perform \emph{next-state prediction}: learning a function that predicts future observations given the current state and an action~\cite{BrysonHo69,ha2018world,sutton}. An environment evolves through latent states $s_t$, which yield observations $o_t$ after actions $a_t$. In the partially observable case, $o_t$ gives only partial information about $s_t$, requiring integration of past history to form a belief state. In the fully observable case, $s_t \approx o_t$, yet predicting in observation space remains difficult because simple physical transitions (e.g., translation) correspond to complex, non-linear changes in high-dimensional pixel space. This necessitates a learned representation space that can linearize these dynamics.  

Our objective is to learn a world model $M$ that can faithfully clone a deterministic environment from purely observational data. The model is trained on a dataset $\mathcal{D}$ consisting of trajectories $\tau = \{(o_1, a_1), \dots, (o_T, a_T)\}$, where $o_t$ is an observation (image) and $a_t$ is an action at timestep $t$.  
The deterministic environment assumption states that, for any initial state and sequence of actions, there is exactly one resulting sequence of observations.
The fidelity of the learned model is assessed through its ability to generate long-horizon rollouts. Specifically, given a starting observation $o_1$ and an action sequence $\{a_t\}_{t=1}^T$, the model produces a rollout $\{\hat{o}_t\}_{t=1}^T$. 
At each timestep $t$, we compute the frame-wise Mean Squared Error (MSE)\footnote{We use frame-wise MSE to measure prediction error because it directly reflects pixel-wise differences and allows straightforward visualization of error accumulation over time. Since the images are normalized in [0,1], MSE can be easily converted to PSNR if needed.} between the generated and ground-truth observations:  
\(
\text{MSE}(t) \;=\; \| o_t - \hat{o}_t \|_2^2.
\)
This produces an error curve $\{\text{MSE}(t)\}_{t=1}^T$ that reveals how prediction error accumulates over time.

\paragraph{Latent Generative Models.} We adopt a latent generative framework~\cite{seo2023masked,high,ha2018world} consisting of two stages: (1) an autoencoder that learns a compressed latent representation~\citep{kingma2013auto, higgins2017beta,vqvae}, and (2) a generative model that captures the dynamics in that latent space~\citep{ha2018world, hafner2019learning, bruce2024genie}. Our work focuses on the representation stage, as its quality critically determines the downstream rollout performance.

\section{Representation as the Bottleneck} \label{oracle}

\begin{figure}[h]
    \centering
    \includegraphics[width=\linewidth]{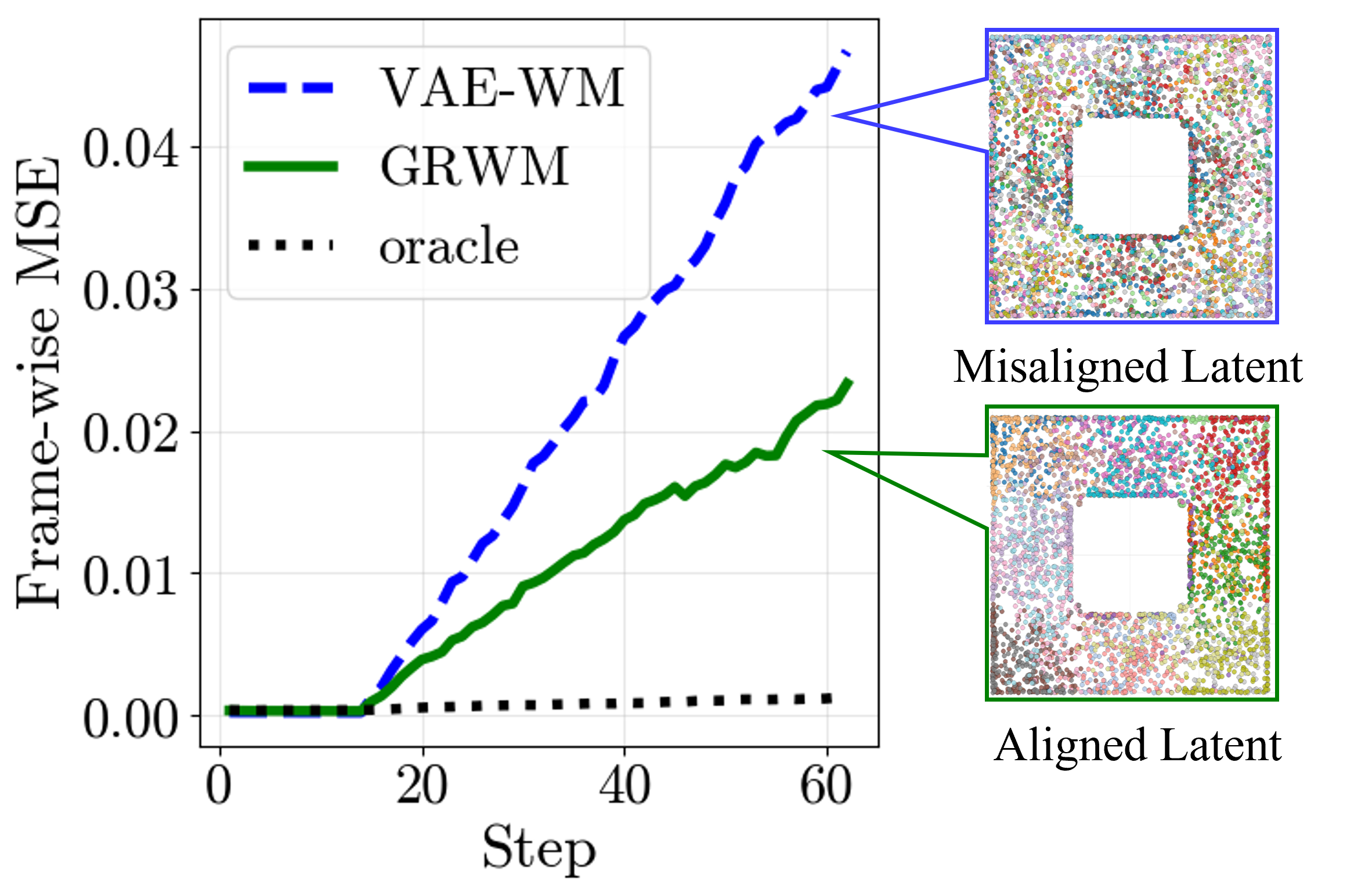}

    \caption{\small \textbf{Representation quality is the primary bottleneck for world model fidelity.}
Frame-wise MSE on a simple deterministic 3D navigation environment.
\textbf{(Left)} An oracle model using ground-truth states (black dotted) achieves near-zero error, establishing a performance upper bound. In contrast, a standard VAE-based world model (blue dashed) accumulates error rapidly.
Our GRWM (green solid) significantly closes this gap by learning a more structurally aligned latent space \textbf{(Right Bottom)}, while the VAE's representation remains disorganized \textbf{(Right Top)}. For further details, see Section~\ref{exp}.}
\label{fig:motivation}
\end{figure}

We begin with a diagnostic question: when a world model fails over long horizons, is the main limitation the dynamics model, or the representation on which that dynamics model operates? In this section, we use the term \emph{bottleneck} to refer to the component that primarily limits rollout fidelity.

This question is especially important in visual world modeling. Even in a simple deterministic environment, the underlying state may be low-dimensional, while the model only observes high-dimensional pixel inputs. In that case, the model must first recover a useful state representation from images before it can predict future observations accurately. If this representation does not preserve the structure of the true state space, then small prediction errors can quickly accumulate during rollout.

To isolate the role of representation, we construct an ``oracle'' world model that bypasses the perception problem and operates directly on the ground-truth state variables, here given by the agent pose $(x,y,\theta)$. The oracle contains three components: (1) an encoder that maps a history of observations to the previous state:
\(
\hat{s}_{t-1} = f_{\text{enc}}(o_{t-k:t-1}),
\)
(2) a dynamics model that predicts the current state from the previous state and action:
\(
\hat{s}_t = f_{\text{dyn}}(\hat{s}_{t-1}, a_t),
\)
and (3) a decoder that reconstructs the current observation from the predicted state:
\(
\hat{o}_t = f_{\text{dec}}(\hat{s}_t).
\)

Figure~\ref{fig:motivation} (left) shows the key result. The plotted MSE is \emph{pixel-space} frame-wise error between predicted and ground-truth observations. Under this evaluation, the oracle maintains near-zero error over long horizons, while VAE-WM accumulates error rapidly. Since the two models are architecturally identical and differ only in the source of the latent state, this comparison leads to two conclusions: first, high-fidelity long-horizon cloning is feasible in this setting; second, the main limitation is not the dynamics model itself, but the representation on which it operates.

Why does a standard latent representation fail? A likely reason is geometric mismatch. A latent space trained mainly for reconstruction does not necessarily preserve the neighborhood structure or global layout of the true state space. As a result, points that are nearby in the true dynamics may not be nearby in latent space, and vice versa. This makes even simple dynamics harder to model reliably over long horizons.

Figure~\ref{fig:motivation} (right) provides an intuitive illustration of this issue. The latent space learned by a standard VAE appears poorly organized with respect to the underlying state structure. This observation motivates the next section, where we introduce geometric regularization to encourage representations that are better aligned with the environment's state manifold.

\section{Geometrically-Regularized World Models}

We next explore how to reduce the gap to oracle-level performance through unsupervised representation learning. Our approach applies the temporal contrastive learning principle~\cite{cltt3,cltt5,cltt7,klindt2020towards} as a geometric regularizer on the autoencoder of a world model. This regularizer shapes the latent space so that its geometry better matches the underlying state manifold, providing a strong inductive bias for stable long-horizon prediction. 

\subsection{Temporal Contextualize Architecture}

Learning underlying states from a single frame is difficult due to \textit{perceptual aliasing}~\cite{alias1,alias2,alias3,alias4}, where distinct states can yield nearly identical observations. For example, two different positions in a maze may appear visually indistinguishable. This also clarifies a key difference in the oracle experiment: the oracle uses coordinates, which encode the global state directly, whereas images provide only local information. A single image cannot specify the global position in environments with heavy aliasing. To recover the true state, the model must aggregate information over time, since a sequence of observations can form a unique signature of the underlying global state~\cite{cltt3,cltt5,cltt7}.

We design the representation model with a causal encoder $E$ and an instantaneous decoder $D$. The encoder maps a sequence of recent observations $(o_{t-k}, \dots, o_t)$ to a latent representation $z_t$, which summarizes the information necessary to infer the current state. The decoder $D$ then reconstructs only the current observation $\hat{o}_t$ from $z_t$:
\[z_t = E(o_{t-k}, \dots, o_t), \quad \hat{o}_t = D(z_t).
\]
This design ensures that $z_t$ is a compact representation of the present state, enriched by past context. To instantiate this design, we build upon variational autoencoder framework augmented with temporal aggregation.

\paragraph{VAE with Temporal Aggregation.} Each frame is first encoded independently with a 2D CNN. The resulting frame-level features are then aggregated by a causal Transformer~\cite{vaswani2017attention} with a sliding temporal window, ensuring that $z_t$ only attends to a limited range of past frames up to and including time $t$. This windowed design captures short-term temporal context while maintaining causality.

\subsection{Temporal Contrastive Regularization}

While the causal encoder introduces temporal context, the standard reconstruction objective on the final frame is insufficient to guarantee a well-structured latent space. The model might learn a ``lazy" solution, ignoring the context and relying solely on the last frame. 
Inspired by principles from contrastive representation learning, we introduce a temporal contrastive regularization loss to explicitly regularize the latent space.

The output of the encoder, a sequence of latent vectors \( \mathbf{z} \in \mathbb{R}^{B \times L \times \dots} \), is first passed through a projection layer (a linear layer) to produce embeddings \( \mathbf{p} \in \mathbb{R}^{B \times L \times D} \)\footnote{\(B\) denotes the batch size, \(L\) the sequence length, and \(D\) the embedding dimension.}~\citep{chen2020simple}.
These embeddings are then $L_2$-normalized to lie on the unit hypersphere:
\(
\mathbf{p}' = \frac{\mathbf{p}}{\|\mathbf{p}\|_2}
\) \citep{wang2020understanding}.
Our regularization objectives are applied to these normalized embeddings, \( \mathbf{p}' \).

\paragraph{Temporal Slowness Loss ($\gL_{\text{slow}}$).} The idea of temporal slowness is that consecutive or nearby states in a trajectory should have similar latent representations, reflecting the intuition that the underlying state of the environment evolves gradually over time~\citep{wiskott2002slow}. Our loss encourages all pairs of frames within the same trajectory's context window to be close to one another on the hypersphere. This enforces that the entire trajectory segment is mapped to a compact and continuous path in the representation space, ensuring that the latent representation evolves slowly and smoothly over time. We formalize this by minimizing the average L2 distance between all pairs of embeddings within a trajectory:
\[
\mathcal{L}_{\text{slow}} = \mathbb{E}_{b \sim \gD} \left[ \mathbb{E}_{(\mathbf{p}'_i, \mathbf{p}'_j) \sim \gP'_b \times \gP'_b} \left[ \left\| \mathbf{p}'_i - \mathbf{p}'_j \right\|_2 \right] \right],
\]
where \( P'_b = \{ \mathbf{p}'_{b,t} \}_{t=0}^{L-1} \) is the set of normalized embeddings for a trajectory \( b \).

\paragraph{Latent Uniformity Loss ($\gL_{\text{uniform}}$).} Slowness alone can lead to feature collapse (i.e., the model maps many inputs to a small region of the latent space). The uniformity loss mitigates this issue by encouraging embeddings to distribute evenly on the hypersphere. It is formally expressed as:

\[
\gL_{\text{uniform}} = \log \mathbb{E}_{(\mathbf{p}'_i, \mathbf{p}'_j) \sim \gP_{\text{neg}}} \left[ e^{-2 \left\| \mathbf{p}'_i - \mathbf{p}'_j \right\|_2^2} \right],
\]

where \( \gP_{\text{neg}} \) is the distribution of all pairs of embeddings from different trajectories in the batch.

\paragraph{Overall Training Objective.}
The complete autoencoder is trained end-to-end by minimizing a total objective function that combines the reconstruction loss, a KL-divergence term from the VAE framework, and our two proposed regularization terms. The final loss is:
\[
\gL_{\text{total}} = \gL_{\text{recon}} + \beta \gL_{\text{KL}} + \lambda_{\text{slow}} \gL_{\text{slow}} + \lambda_{\text{uniform}} \gL_{\text{uniform}}
\]
where \( \beta \), \( \lambda_{\text{slow}} \), and \( \lambda_{\text{uniform}} \) are hyperparameters that balance the contribution of each term.

\section{Experiments}\label{exp}
\subsection{Setup}\label{exp:setup}
\begin{figure}[h]
    \centering
    \begin{subfigure}[b]{0.12\textwidth}
        \centering
        \includegraphics[width=\textwidth, trim=13 13 13 13, clip]{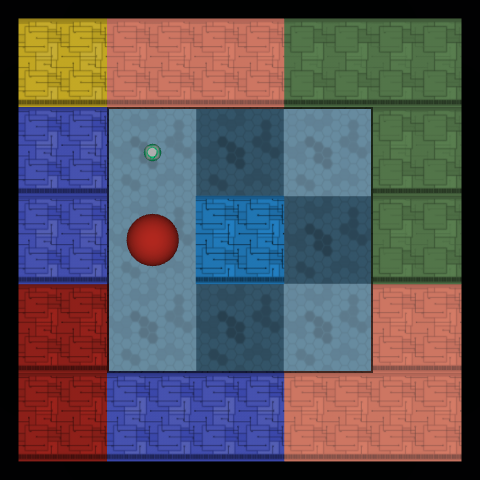}
        \caption{M3$\times$3-DET}
        \label{fig:map_3x3}
    \end{subfigure}
    \begin{subfigure}[b]{0.12\textwidth}
        \centering
        \includegraphics[width=\textwidth, trim=13 13 13 13, clip]{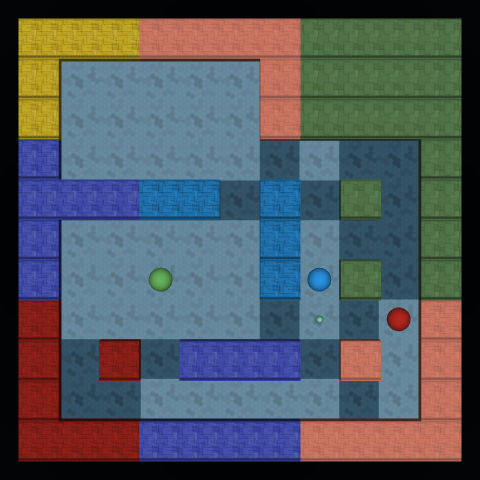}
        \caption{M9$\times$9-DET}
        \label{fig:map_9x9}
    \end{subfigure}
    \begin{subfigure}[b]{0.12\textwidth}
        \centering
        \includegraphics[width=\textwidth, trim=13 13 13 13, clip]{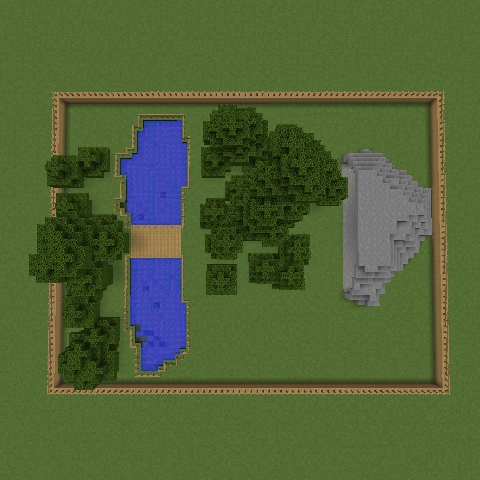}
        \caption{MC-DET}
        \label{fig:map_mc}
    \end{subfigure}
    \caption{\small Top-down visualizations of our three closed environments: M3$\times$3-DET, M9$\times$9-DET, and MC-DET. These maps illustrate the overall layout and are for visualization purposes only; they are not provided as input to the agent. The agent's input is restricted to first-person observations. For a more representative depiction of the agent's surroundings, high-angle perspective views are also included in the supplementary material, offering a better sense of the environments' three-dimensional structure and scale.}\vspace{-5mm}
    \label{fig:maps}
\end{figure}
\paragraph{Datasets.} We introduce three datasets collected in deterministic environments~\cite{gornet2024automated,pasukonis2022memmaze,johnson2016the}: Maze 3$\times$3-DETERMINISTIC (M3$\times$3-DET), Maze 9$\times$9-DETERMINISTIC (M9$\times$9-DET), and Minecraft-DETERMINISTIC (MC-DET). Trajectories in these datasets are sequences of (action, observation) pairs from a first-person perspective. The map layout for each environment is fixed, rendering the trajectories fully deterministic. Figure~\ref{fig:maps} shows top-down views of these environments for visualization; these maps are not available to the agent. The Maze datasets differ in size and complexity, while the Minecraft dataset provides richer visual observations. Further details on data collection are in the supplementary material. 

\paragraph{Baselines.} 
State-of-the-art latent generative world models generally adopt a standard VAE as the backbone for representation learning. Our method does not modify the dynamics model itself. Therefore, when comparing with existing sota approaches, we pair our representation module with the same dynamics models they use.
We select three sota latent generative world models: Standard Diffusion (SD)~\citep{alonso2024diffusionworldmodelingvisual}, Video Diffusion Models (VD)~\citep{ho2022video}, and Diffusion Forcing (DF) ~\citep{chen2024diffusion}.
A detailed description of training configurations is provided in the supplementary material. 

\paragraph{Metrics.} 
We evaluate prediction fidelity using \textit{frame-wise MSE}. Since the environments are deterministic, a unique ground-truth future exists for any given initial state and action sequence. We can therefore compute the mean squared error between the predicted and ground-truth observations at each timestep in pixel space. This metric reflects the step-by-step discrepancy between predicted and actual observations across the entire trajectory. A lower frame-wise MSE indicates higher fidelity in cloning the environment's dynamics over time.
\subsection{Rollout Fidelity}

\begin{figure*}[h]
    \centering
    \begin{subfigure}[b]{0.33\textwidth}
        \centering
        \includegraphics[width=\textwidth]{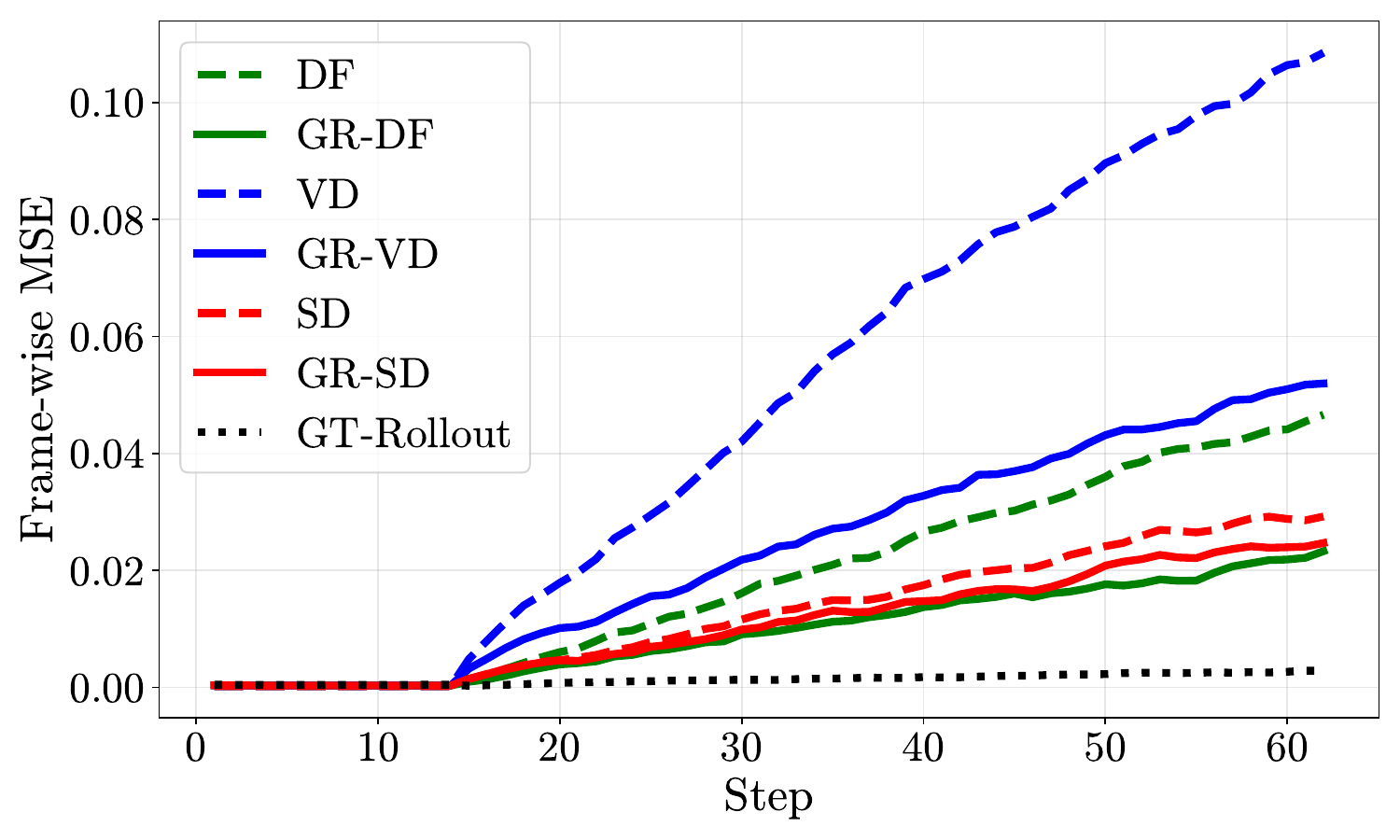}
        \caption{M3$\times$3-DET}
        \label{fig:rollout_3x3}
    \end{subfigure}
    \hfill
    \begin{subfigure}[b]{0.33\textwidth}
        \centering
        \includegraphics[width=\textwidth]{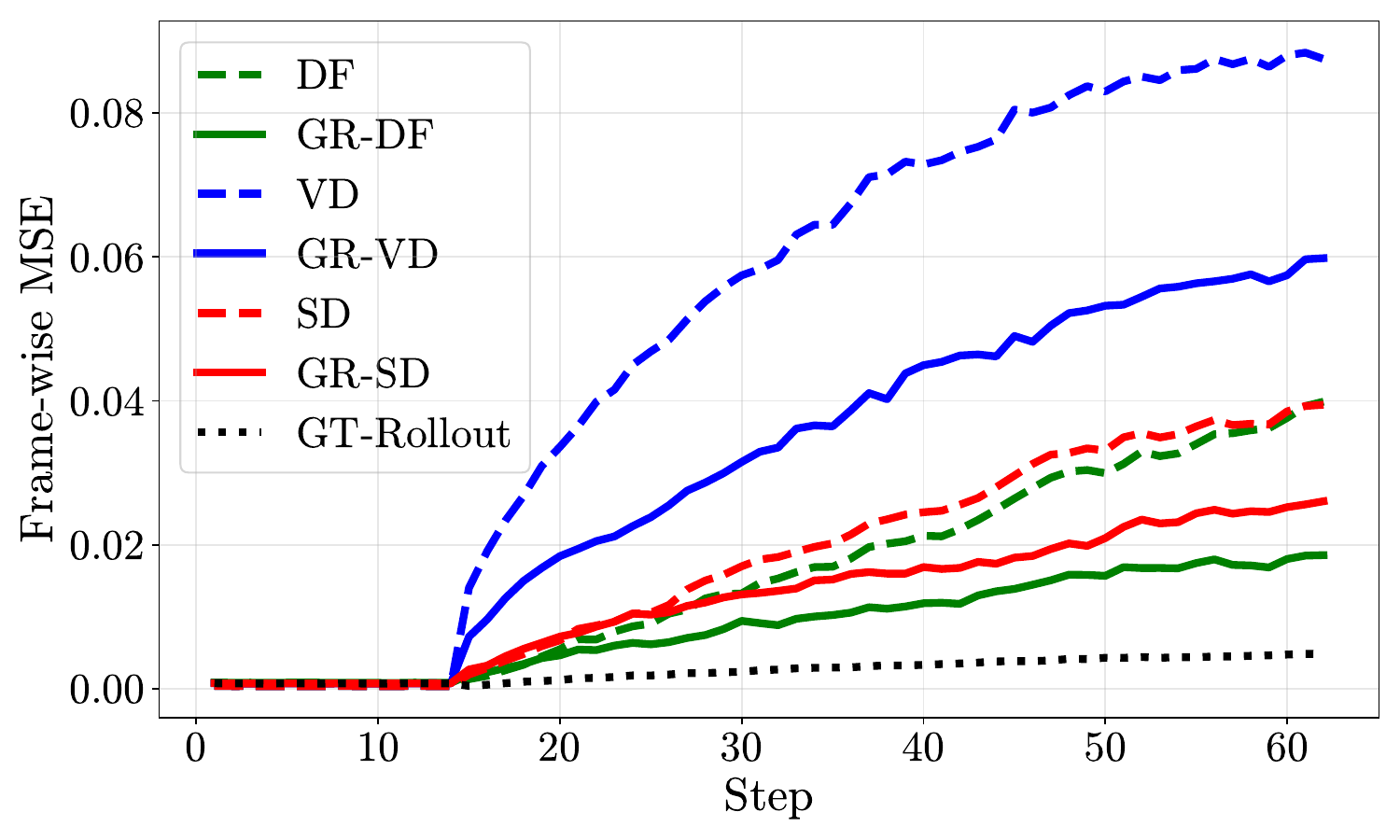}
        \caption{M9$\times$9-DET}
        \label{fig:rollout_9x9}
    \end{subfigure}
    \hfill
    \begin{subfigure}[b]{0.33\textwidth}
        \centering
        \includegraphics[width=\textwidth]{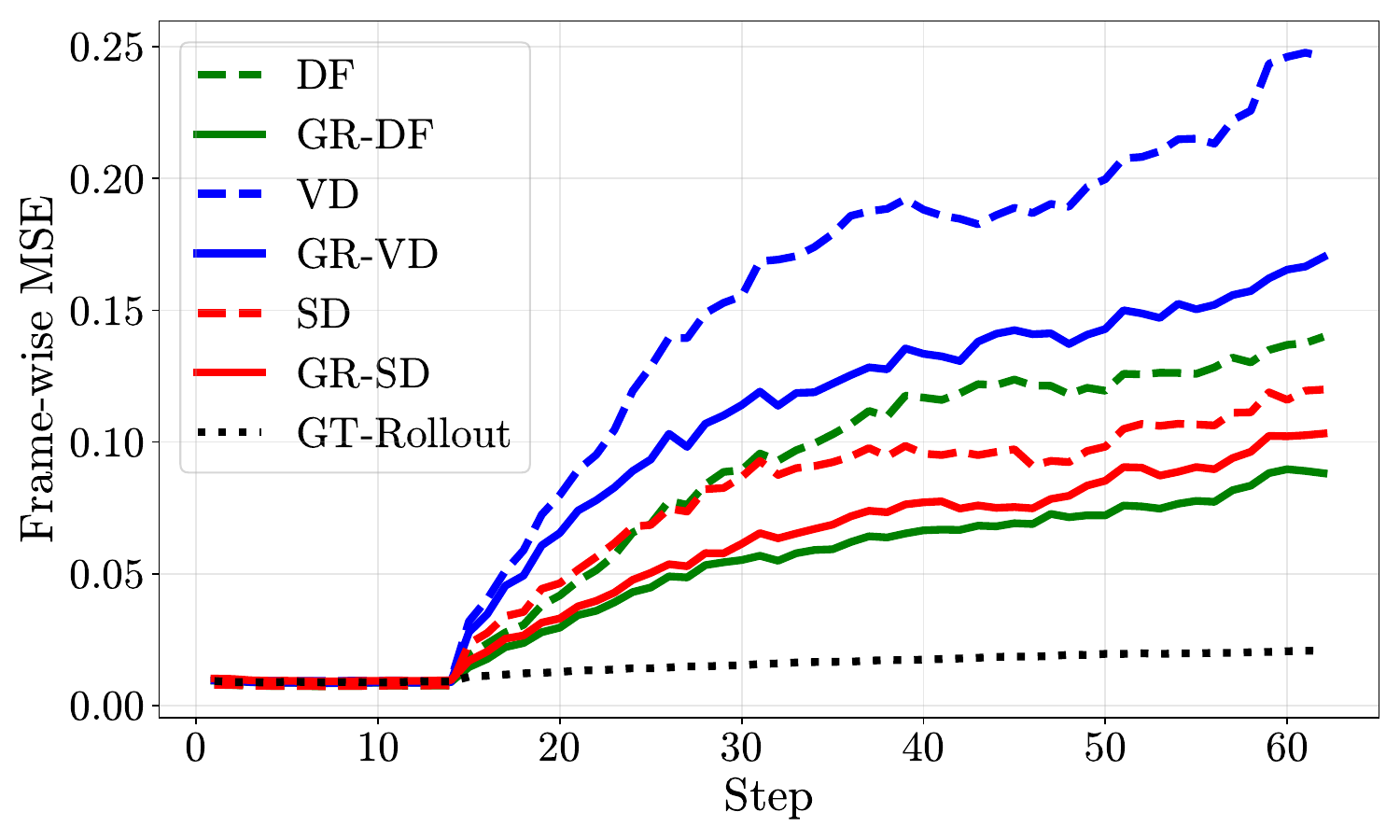}
        \caption{MC-DET}
        \label{fig:rollout_mc}
    \end{subfigure}
    \caption{\small Rollout Performance. Frame-wise MSE between predicted and ground-truth trajectories on (a) M3x3-DET, (b) M9x9-DET, and (c) MC-DET datasets. The oracle model (black dotted line), which operates on the true underlying states, establishes a lower bound on error. For all three dynamics models—Diffusion Forcing (DF), Video Diffusion (VD), and Standard Diffusion (SD)—our GRWM (solid lines) consistently outperforms baselines (dashed lines), demonstrating significantly lower error accumulation over 63 steps and substantially closing the performance gap to the oracle.}
    \label{fig:rollout_mse}
\end{figure*}

To quantitatively evaluate rollout prediction fidelity, we present our main results in Figure~\ref{fig:rollout_mse}. The results demonstrate a consistent advantage for GRWM, and the performance gap widens with rollout length. GRWM (solid lines) maintains a significantly lower prediction error compared to the baseline using a vanilla VAE (dashed lines). Baseline models accumulate error quickly, causing trajectories to diverge, whereas our method maintains a much flatter error curve. This shows stronger long-term temporal consistency.

\subsection{Qualitative Long-Horizon Results}

To further probe the long-term stability, we conduct an extreme long-horizon generation task. For this experiment, we specifically select sota world model, Diffusion Forcing, and the M9x9-DET and MC-DET dataset. In the following discussion, we refer to the original Diffusion Forcing as \textbf{VAE-WM}, and GRWM with Diffusion Forcing as \textbf{GRWM}. 
We use a sequence of randomly sampled actions to simulate an trajectory.

\begin{figure}[h]
\centering
\includegraphics[width=\linewidth, trim={0 0 3cm 0}, clip]{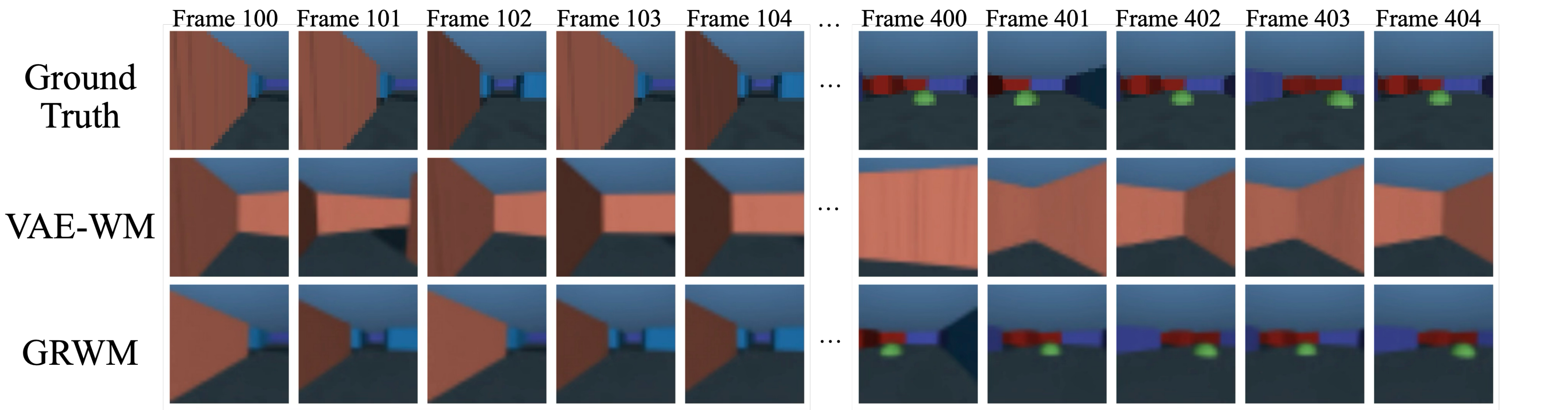}
\caption{\small Qualitative comparison of medium-horizon rollouts in M9x9-DET. We visualize consecutive frames around frame 100 and frame 400. Our method (GRWM) maintains high similarity to the ground truth throughout, while the baseline VAE-WM gets trapped near the pink wall, indicating that VAE-WM tends to  ``teleport" between visually similar but distinct locations.}
\label{fig:medium_horizon}
\end{figure}

\begin{figure}[h]
\centering
\includegraphics[width=\linewidth]{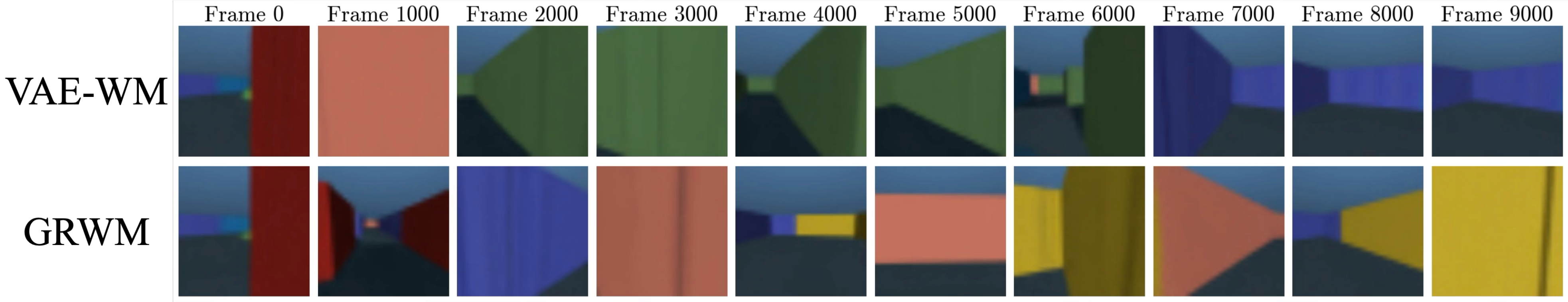}
\caption{\small Qualitative comparison of ultra long-horizon rollouts on the Maze 9x9-CE dataset. Frames are sampled every 1000 steps from a 10,000-step rollout. The baseline VAE-WM frequently gets stuck generating the same color states, failing to explore the environment effectively. In contrast, GRWM produces a coherent and diverse trajectory, successfully exploring different regions while preserving long-term temporal and structural consistency.}\vspace{-5mm}
\label{fig:long_horizon}
\end{figure}

We evaluate our model on trajectories of varying lengths. To test long-horizon stability, we tasked the model with producing 10,000 consecutive frames from a single starting point. We visualize these long trajectories in Figure~\ref{fig:medium_horizon} and Figure~\ref{fig:long_horizon}. For the more complex MC-DET environment, we specifically evaluate a 100-frame  rollout, as shown in Figure~\ref{fig:mc_medium}. Additional results under different starting conditions are provided in the supplementary material. 

\begin{figure}[h]
\centering
\includegraphics[width=\linewidth]{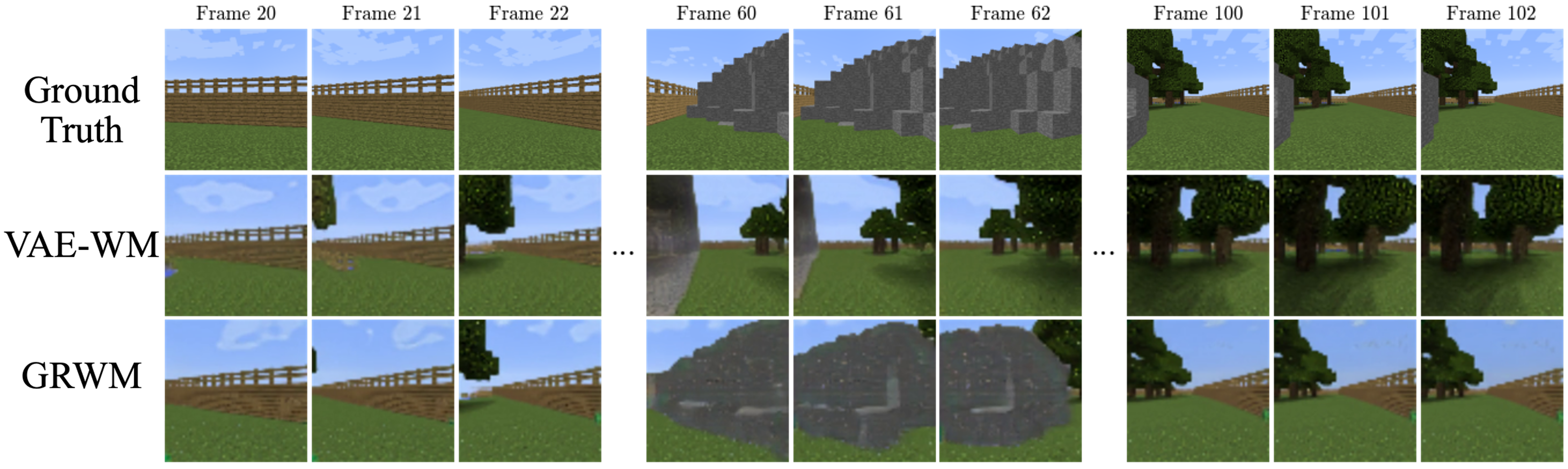}
\caption{\small Qualitative comparison of medium-horizon rollouts in MC-DET. We visualize rollouts from a baseline VAE-based world model (VAE-WM, middle) and our method (GRWM, bottom) against the ground truth (top). The baseline VAE-WM fails to model the complex camera trajectory, diverging significantly and rendering incorrect objects (e.g., trees instead of the stone wall at frame 60). Our method (GRWM) successfully tracks the complex motion and maintains high-fidelity generation consistent with the ground truth throughout the sequence.}
\vspace{-4mm}
\label{fig:mc_medium}
\end{figure}

The results reveal a critical failure mode in the baseline model. As seen in the figures, the VAE-WM quickly succumbs to mode collapse, getting trapped in repetitive loops that render nearly identical, low-complexity frames. Our interpretation is that the model learns to  ``teleport" between visually similar but causally disconnected regions of the environment. For example, it can spend thousands of consecutive frames generating views of a single-colored wall (e.g., the pink wall in Figure~\ref{fig:medium_horizon}; the green and blue walls in Figure~\ref{fig:long_horizon}). 
The pixel-based reconstruction loss forces the model to map visually similar observations—such as different walls of the same color—to nearby points in the latent space, irrespective of their true distance or causal connection within the environment. This creates   ``attractor states" and an entangled manifold. The dynamics model, operating in this flawed space, learns that jumping between these close latent points is a low-cost action, resulting in the observed ``teleportation" between safe havens of low reconstruction error instead of navigating the true, complex topology of the world.

In contrast, GRWM generates a far more coherent and diverse trajectory because its representation is regularized to be consistent with the true underlying state manifold. As shown in Figure~\ref{fig:medium_horizon}, GRWM maintains high fidelity at both 100 and 400 frames, while the VAE-WM fails. In the longer rollout (Figure~\ref{fig:long_horizon}), the sequence of frames from GRWM clearly shows movement and exploration. The appearance of the yellow wall in later frames is not a random plausible image; it is evidence of a continuous traversal through a latent space that mirrors a physically possible path in the environment. Our geometric regularization forces the model to respect the world's structure, preventing the state-skipping and teleportation artifacts that dominate the VAE rollouts.

\subsection{Latent Representation Analysis}

\paragraph{Latent Probing.}
We perform a latent probing analysis to quantitatively assess how well the learned representations capture the true underlying states of the environment (i.e., agent position $(x,y)$ and orientation $\theta$). We freeze the trained autoencoder and use its encoder to obtain latent vectors for all observations. A small MLP probe is then trained to predict the ground-truth states from these latent vectors. We report regression MSE on a held-out validation set, where a lower value indicates that the latent space is more informative and better aligned with the environment’s true state manifold.

\begin{table}[ht]
\centering
\caption{\small Latent probing analysis. GRWM consistently learns representations that are more predictive of the true underlying states. We report regression MSE of an MLP probe on a held-out set (lower is better).}
\label{tab:latent_probing}
\begin{tabular}{lccc}
\toprule
\textbf{Model} & \textbf{M3$\times$3-DET} & \textbf{M9$\times$9-DET} & \textbf{MC-DET} \\
\midrule
VAE-WM & 0.082 & 0.106 & 0.137 \\
GRWM & 0.031 & 0.058 & 0.081 \\
\bottomrule
\end{tabular}
\end{table}

\vspace{-2mm}
GRWM consistently learns latent representations that are more linearly predictive of the ground-truth agent state. The results, summarized in Table~\ref{tab:latent_probing}, clearly support our hypothesis. Across all three datasets, our method leads to a significant reduction in regression MSE. Notably, the improvement is consistent regardless of the environment's complexity, highlighting the general applicability and effectiveness of our approach in structuring the latent space.

\begin{figure}[h]
  \centering
  \begin{tabular}{ccc}
    \textbf{M3x3} & \textbf{M9x9} & \textbf{MC} \\
    \includegraphics[width=0.28\linewidth,trim={350 350 0 0},clip]{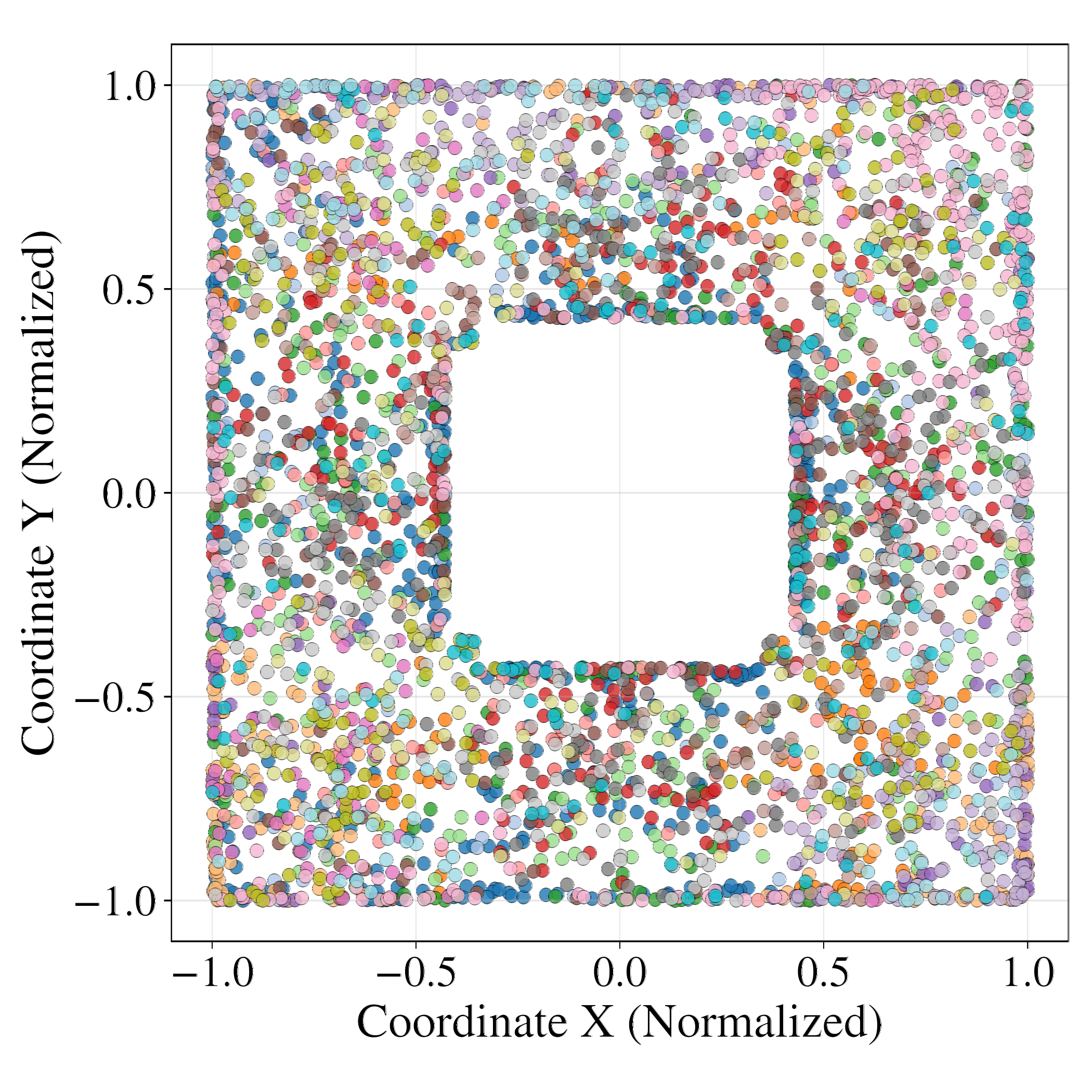} &
    \includegraphics[width=0.28\linewidth,trim={350 350 0 0},clip]{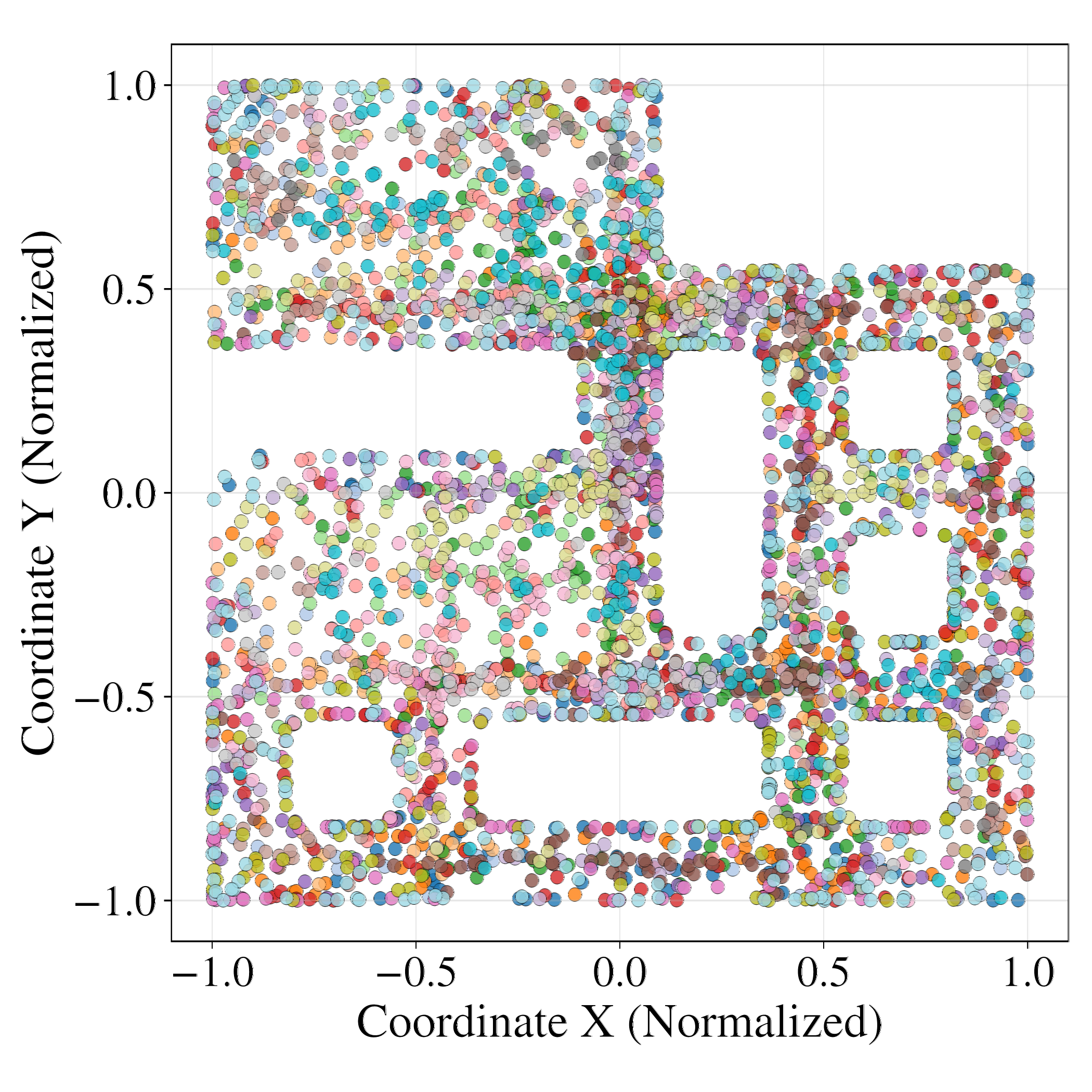} &
    \includegraphics[width=0.28\linewidth,trim={350 350 0 0},clip]{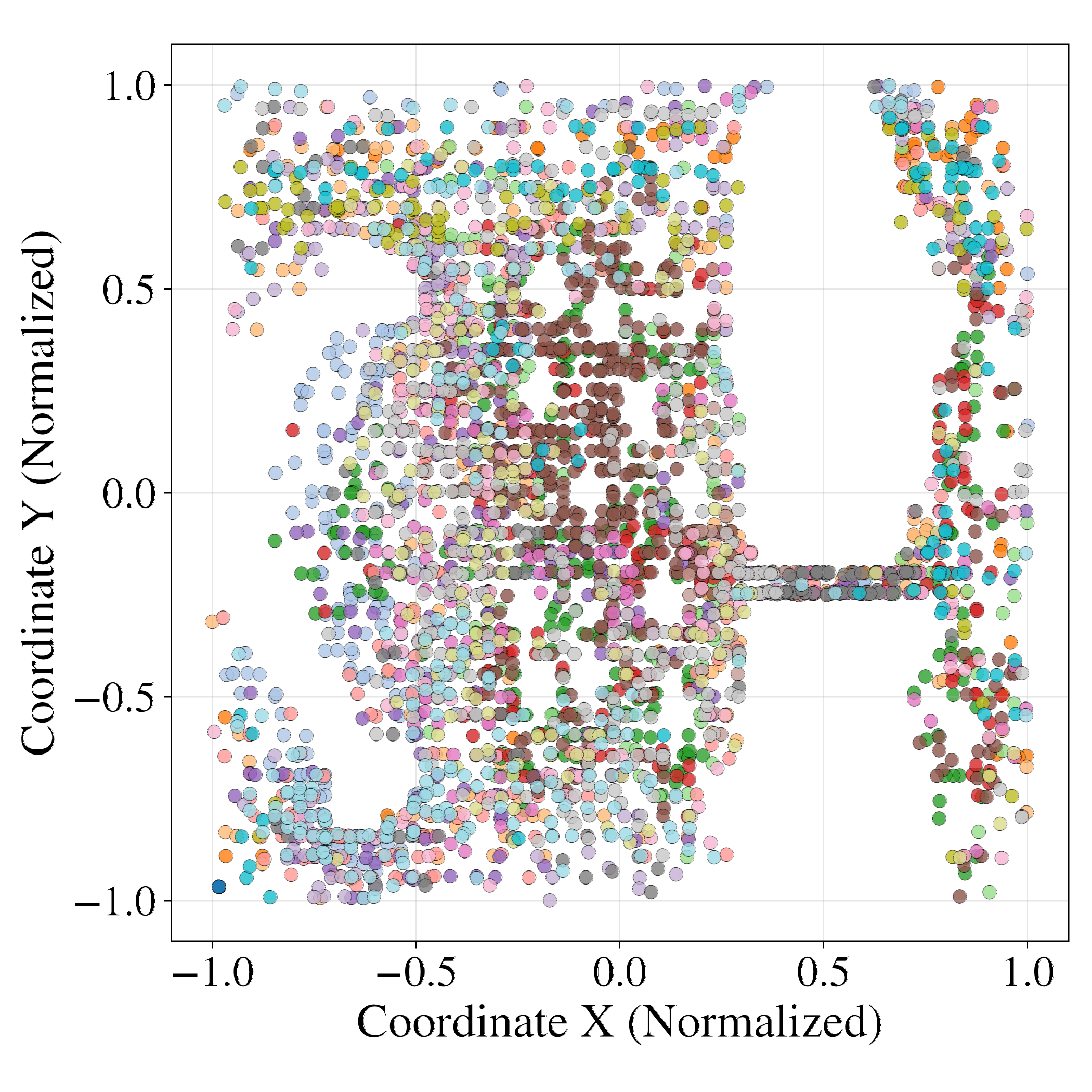} \\
    VAE-WM & VAE-WM & VAE-WM \\
    \includegraphics[width=0.28\linewidth,trim={350 350 0 0},clip]{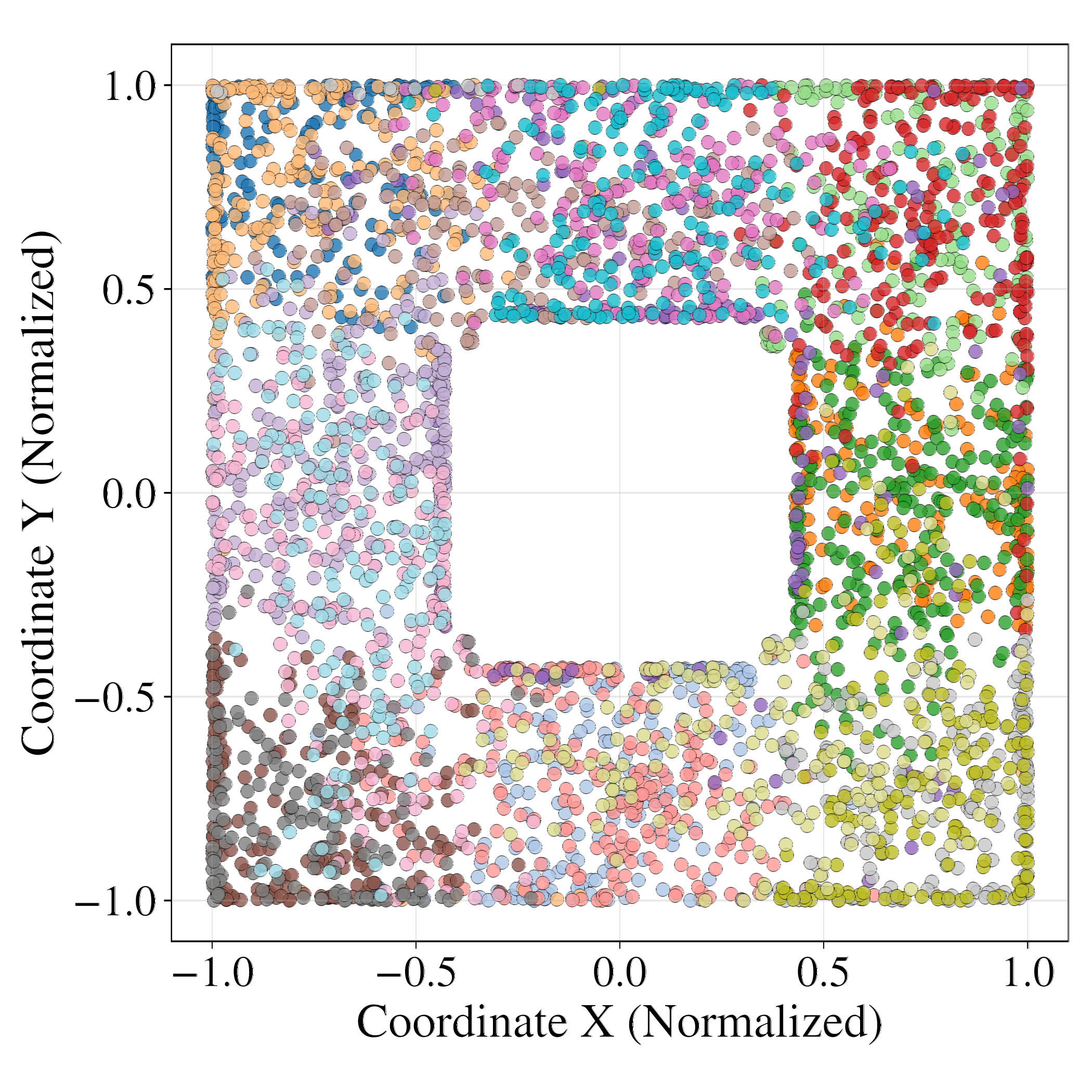} &
    \includegraphics[width=0.28\linewidth,trim={350 350 0 0},clip]{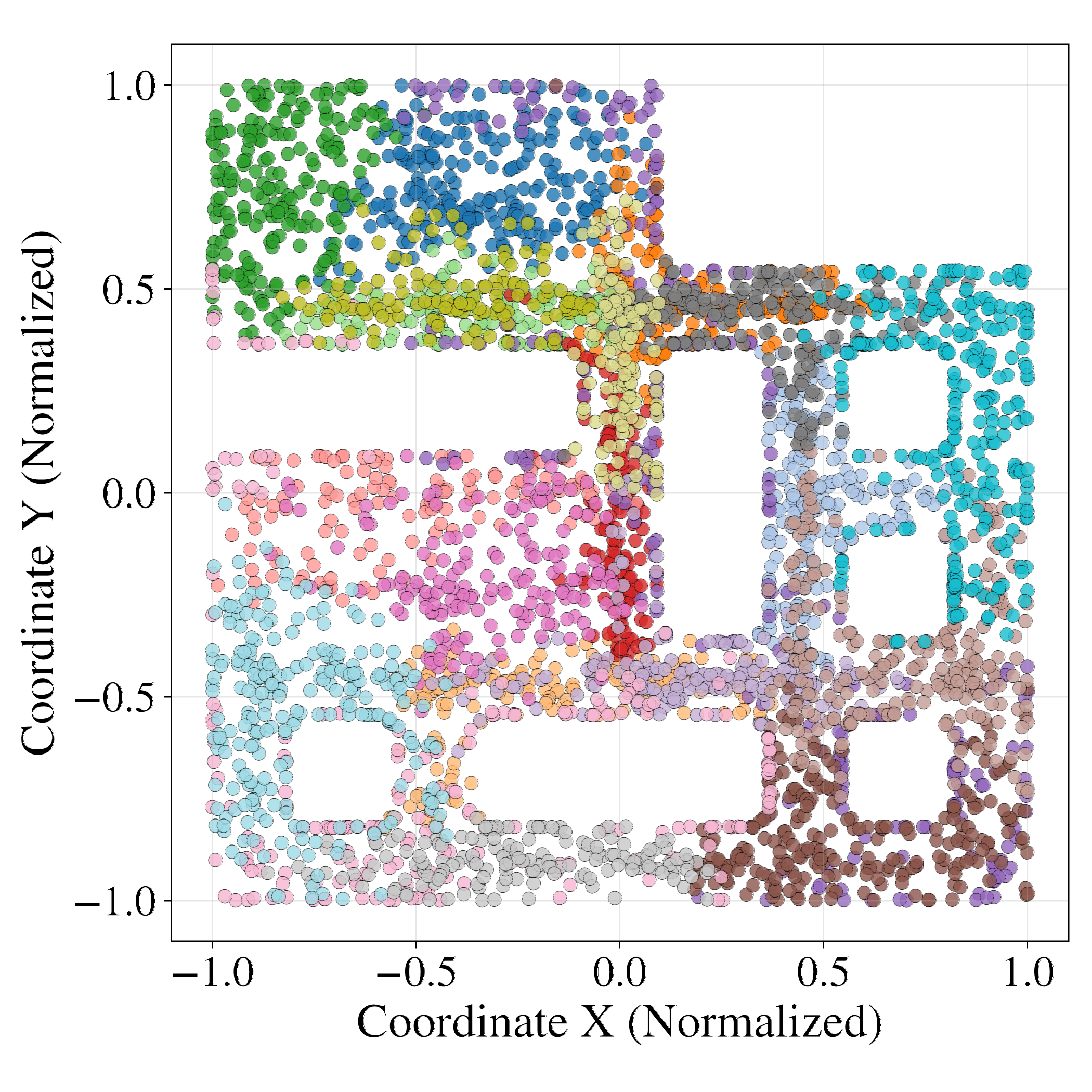} &
    \includegraphics[width=0.28\linewidth,trim={350 350 0 0},clip]{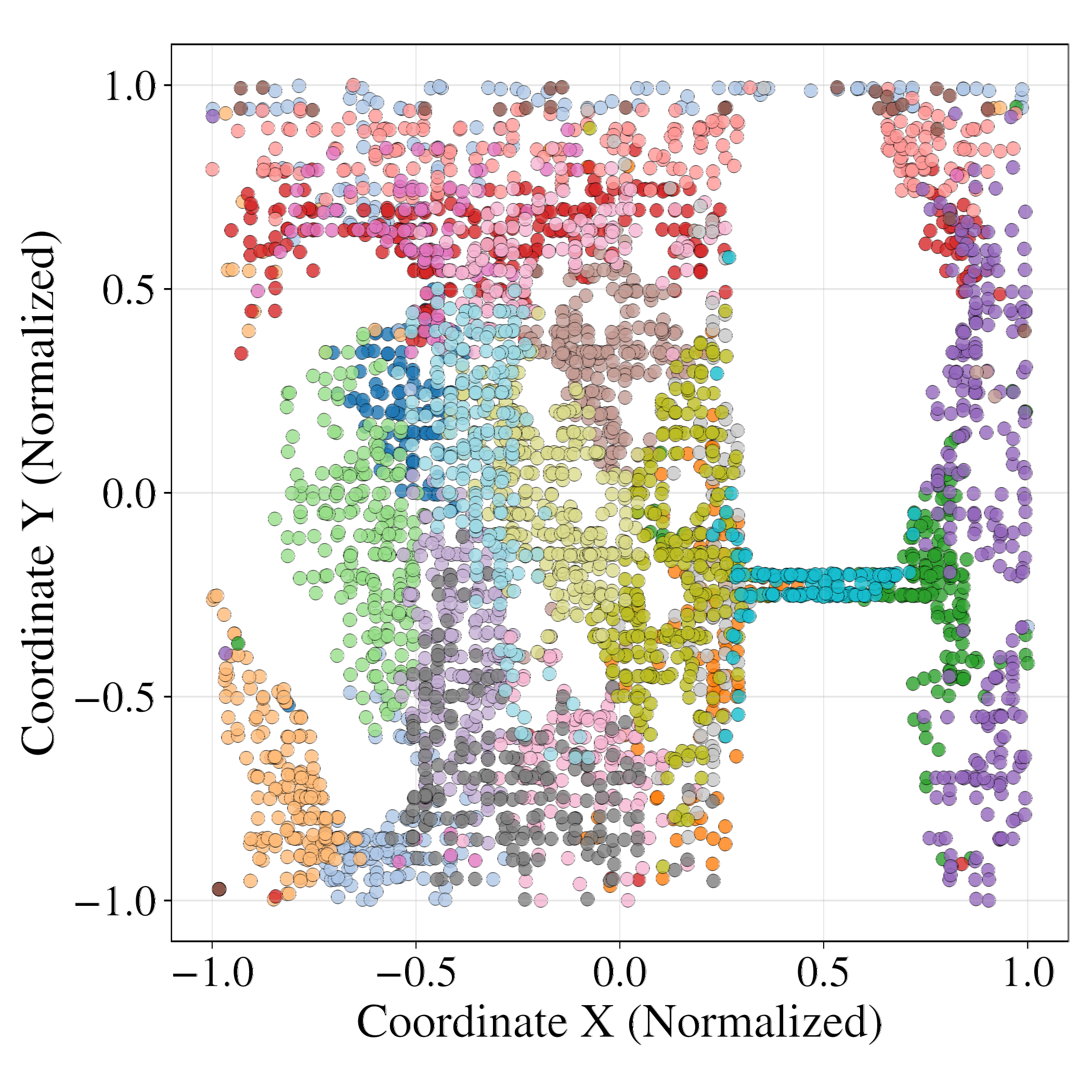} \\
    GRWM & GRWM & GRWM \\
\end{tabular}

\caption{\small Visualization of latent space structure through clustering analysis. 
We perform $k$-means clustering ($k=20$) on the latent representations of frames. Each point in the plots corresponds to a frame, positioned according to its true ($x,y$) coordinates in the environment. The $(x,y)$ coordinates are normalized and lie within $[-1,1]$. Points are colored based on their assigned latent cluster ID. The top row (VAE-WM) shows scattered, noisy clusters, indicating that spatially distant frames are incorrectly grouped together. The bottom row (GRWM) shows well-defined, spatially coherent clusters, demonstrating that our learned latent space is structurally aligned with the environment's true state manifold.}
\vspace{-5mm}
\label{fig:latent_clusters}
\end{figure}

\paragraph{Latent Clustering.}
To further investigate the structure of the learned representations, we conduct a clustering analysis. We first obtain latent vectors for a set of frames using the trained encoder and then apply a k-means algorithm (with $k=20$ clusters) to group these vectors in the latent space. To visualize the result, we plot each frame as a point at its ground-truth $(x,y)$ position within the environment and assign it a color based on its latent cluster ID. The results are shown in Figure~\ref{fig:latent_clusters}.

GRWM successfully forces the model to learn a latent manifold that is structurally aligned with the environment's true topology. The baseline VAE (top row) produces noisy and fragmented clusters. A single color (representing a single latent cluster) is scattered across disparate and often distant regions of the map. This indicates a highly entangled representation, where frames from fundamentally different underlying states are incorrectly mapped to the same region of the latent space. Such a representation provides a poor foundation for a dynamics model, as it fails to distinguish between causally distinct states. In contrast, our method (bottom row) produces remarkably coherent and spatially contiguous clusters. Each color largely corresponds to a distinct, localized region of the environment, such as a specific corridor or room. By correctly grouping states that are close in the physical world, our method provides a smooth and well-structured landscape upon which a dynamics model can learn accurate and generalizable transitions.

\section{Ablation Studies}

We conduct ablation studies to validate the contribution of our core components and design choices. 
Specifically, we examine four aspects: (1) the necessity of the two core loss terms, (2) the role of the projection head, (3) the impact of latent dimension, and (4) the effect of critical design choices on model performance. 
The detailed results are provided in the supplementary material.

\section{Limitations}
\paragraph{Environment Complexity.} The current evaluation is restricted to relatively simple and controlled environments. Scaling to more complex settings—such as partially observable, stochastic, or photorealistic environments—remains unresolved. These settings introduce additional structure and uncertainty, and require representations that can support consistent long-horizon predictions under such conditions.

\paragraph{Visual Artifacts.} GRWM still produces subtle visual artifacts and fails to recover fine-grained scene details, particularly in more complex environments such as Minecraft-DET. This indicates that the learned latent representation does not fully capture high-frequency or semantically relevant structure. Improving representation quality and decoding fidelity is a clear direction for future work.

\paragraph{Computation Overhead.} GRWM introduces additional computation beyond a standard VAE-based world model due to temporal aggregation and geometric regularization. Since the encoder uses a Transformer over temporal context, the overhead can increase as the sequence length grows.

\paragraph{Gap to Oracle Performance.} Although GRWM substantially narrows the gap to the oracle model, a large performance gap still remains. This suggests that even in deterministic environments, the learned representation is still far from sufficient to recover the full state information needed for near-perfect long-horizon prediction. Closing this gap remains an open problem, and further progress will likely require stronger representation learning methods.

\section{Conclusion}
\vspace{-1mm}
In this work, we studied the problem of building high-fidelity world models for deterministic, closed 3D environments. Our starting point is that the representation quality is crucial. Much prior work has focused on designing stronger dynamics models, but our results show that the main limitation in long-horizon prediction comes from the structure of the latent space in which the dynamics operate in. We apply temporal contrastive principles to shape the latent space in latent generative world model. This allows the dynamics models to reach their full capability. It narrows the gap to the oracle and supports our main hypothesis: representation quality is a key factor for achieving reliable long-horizon prediction.
{
    \small
    \bibliographystyle{ieeenat_fullname}
    \bibliography{main}
}

\clearpage
\onecolumn
\appendix
\section*{\centering \bf Supplementary Material}

\section{Training Details}
\label{app:training}

Detailed training configurations, including hyperparameters and implementation details, are available in the accompanying code repository:
\url{https://github.com/XiaFire/Clone_Deterministic_Environment}.



\section{Additional Evaluation Metrics}
\label{app:metrics}

In deterministic environments, there exists a single ground-truth trajectory for a given initial condition and action sequence. As a result, pixel-wise MSE provides a direct and reliable measure of fidelity, and is used as our primary metric. 

To complement this evaluation, we additionally report perceptual metrics on Maze 9$\times$9. These metrics capture structural similarity and distributional alignment that are not fully reflected by MSE. All metrics are computed in pixel space over rollout sequences. As shown in Table~\ref{exp:metric}, the improvements are consistent with those observed under MSE.

\begin{table}[h]
\centering
\scriptsize
\caption{Supplementary perceptual metrics on Maze 9$\times$9.}
\label{exp:metric}
\begin{tabular}{lcccc}
\toprule
 & \multicolumn{2}{c}{SSIM $\uparrow$} 
 & \multicolumn{2}{c}{rFID $\downarrow$} \\
Model
 & Baseline & GR
 & Baseline & GR \\
\midrule
DF
 & 0.8448 & \cellcolor{gray!20}0.8516
 & 4.4345 & \cellcolor{gray!20}2.8729 \\
VD
 & 0.5979 & \cellcolor{gray!20}0.7537
 & 18.1453 & \cellcolor{gray!20}7.2813 \\
SD
 & 0.8367 & \cellcolor{gray!20}0.8369
 & 6.4686 & \cellcolor{gray!20}4.5200 \\
\bottomrule
\end{tabular}
\vspace{-3mm}
\end{table}

\section{Additional Atari Experiments}
\label{app:atari}

We include additional experiments on Atari. The goal of this experiment is to extend our evaluation beyond static 3D environments to dynamic settings. We report PSNR and SSIM computed on rollout frames in pixel space. As shown in Table~\ref{exp:atari}, GRWM consistently improves over the VAE-based world model across the evaluated environments. We note that this experiment is limited in scale and is not intended as a full Atari benchmark, but rather as a supplementary validation.

\begin{table}[h]
\centering
\scriptsize
\caption{\small Additional Atari results.}
\label{exp:atari}
\begin{tabular}{lcccc}
\toprule
Env.
& \multicolumn{2}{c}{PSNR $\uparrow$}
& \multicolumn{2}{c}{SSIM $\uparrow$} \\
\cmidrule(lr){2-3} \cmidrule(lr){4-5}
& VAE-WM & GRWM & VAE-WM & GRWM \\
\midrule
Asterix
& 28.57 & \cellcolor{gray!20}29.04
& 0.9479 & \cellcolor{gray!20}0.9518 \\
Breakout
& 34.23 & \cellcolor{gray!20}37.76
& 0.9848 & \cellcolor{gray!20}0.9872 \\
\bottomrule
\end{tabular}
\vspace{-4mm}
\end{table}
\section{Dataset Details}\label{sec:appendix_datasets}

We evaluate our models on two environments: a memory Maze environment and a Minecraft environment.  

\paragraph{Maze.} 
For the Maze environment, we fix the random seed to generate a consistent set of maps. We use Memory-Maze Environment~\citep{pasukonis2022memmaze}. The rendered images are obtained using the MuJoCo engine. The agent has a discrete action space consisting of \{move forward, turn left, turn right\}. Trajectories are collected with a noisy A* algorithm to ensure sufficient coverage of the maze.  

\paragraph{Minecraft.} 
For the Minecraft environment, we adopt the map from~\citet{gornet2024automated} and enclose the area with wooden fences to restrict exploration. Trajectories are generated using a noisy A* policy under the same action space as in the Maze environment, providing diverse yet structured coverage. We first record the underlying deterministic state trajectories, and then render the corresponding pixel observations using Blender.

\paragraph{Statistics.} 
Each trajectory contains up to 1000 frames, though most consist of several hundred frames. Each dataset contains 5000 trajectories in total.  

\begin{figure}[h]
    \centering
    \begin{tabular}{ccc}
        \includegraphics[width=0.3\textwidth]{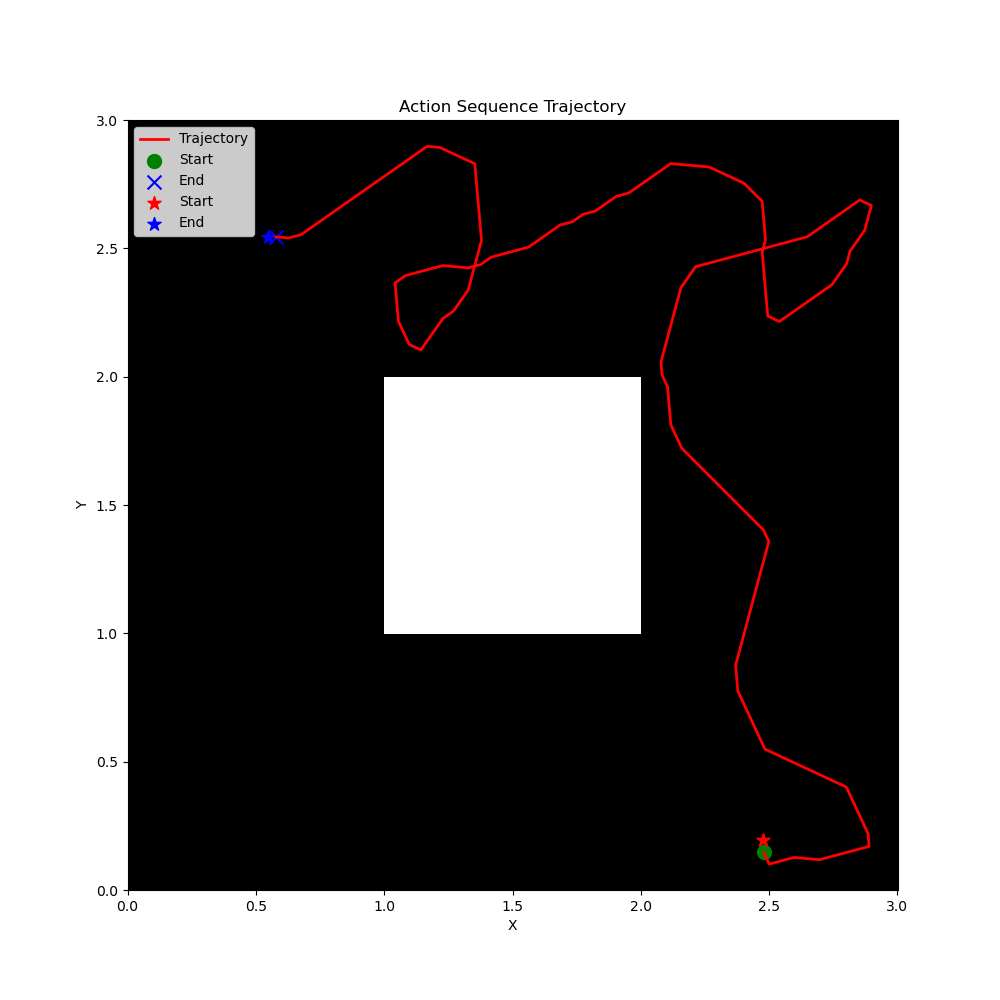} &
        \includegraphics[width=0.3\textwidth]{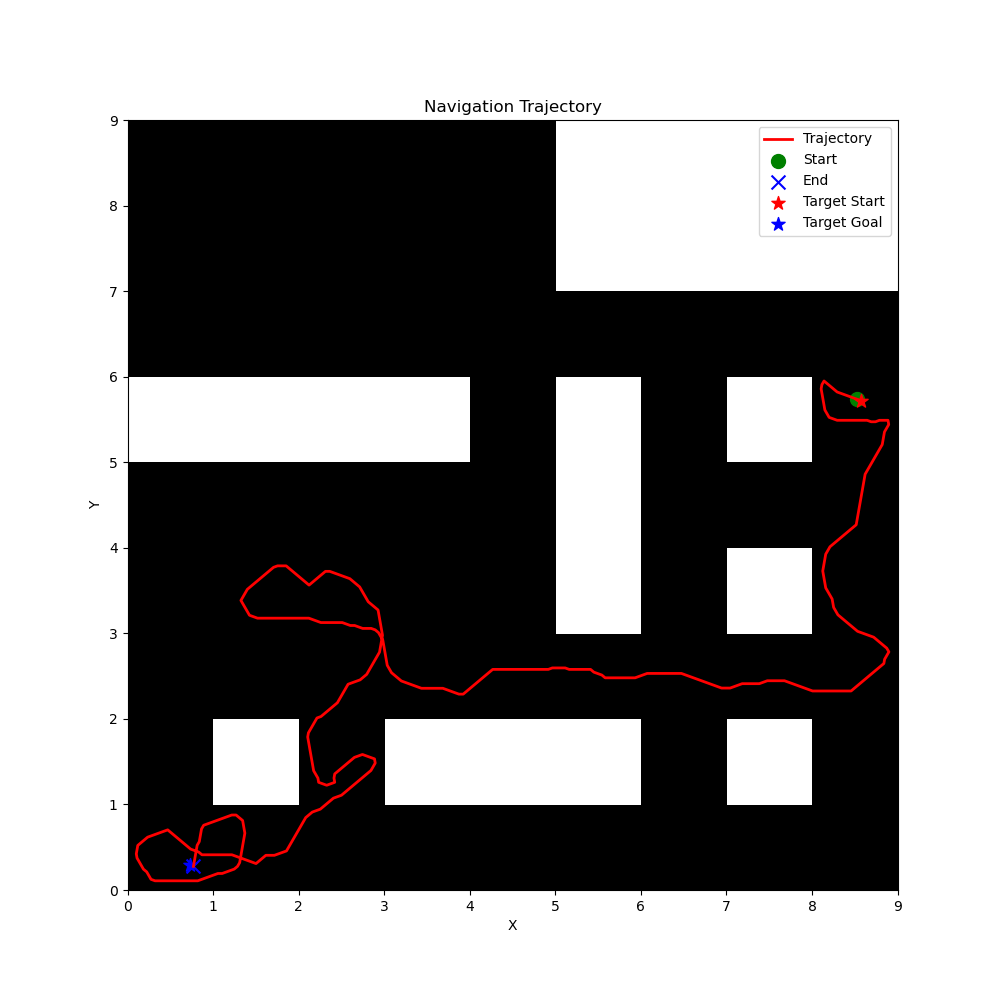} &
        \includegraphics[width=0.3\textwidth]{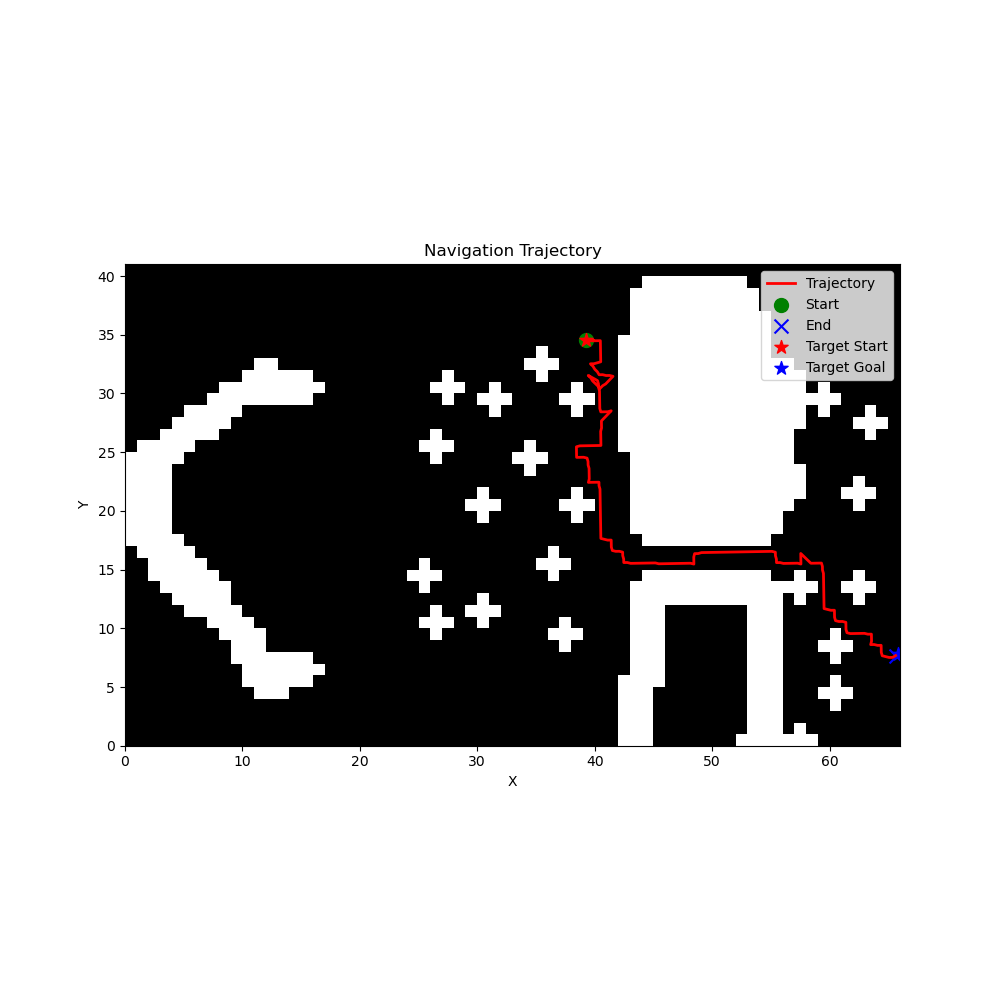} \\
        M3x3-DET & M9x9-DET & MC-DET
    \end{tabular}
    \caption{Representative trajectories from the three datasets. Each plot shows a sample trajectory overlaid on the environment layout.}
    \label{fig:traj_vis}
\end{figure}

\paragraph{Trajectory Visualization.}  
To provide an intuitive understanding of the datasets, we visualize several representative trajectories. 
Figure~\ref{fig:traj_vis} shows examples from three settings: M3x3-DET, M9x9-DET, and MC-DET.  

\begin{figure}[h]
    \centering
    \begin{subfigure}[b]{0.25\textwidth}
        \centering
        \includegraphics[width=\textwidth, trim=13 13 13 13, clip]{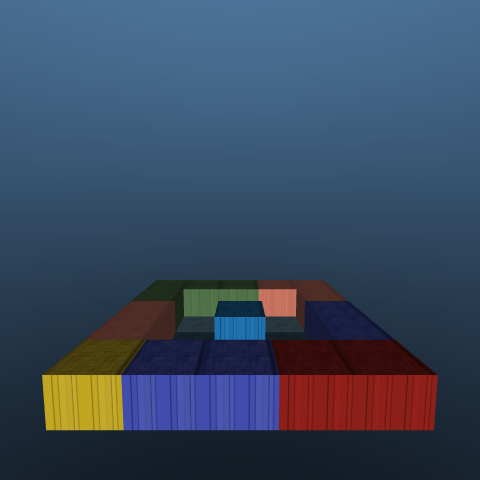}
        \caption{M3$\times$3-DET}
    \end{subfigure}
    \begin{subfigure}[b]{0.25\textwidth}
        \centering
        \includegraphics[width=\textwidth, trim=13 13 13 13, clip]{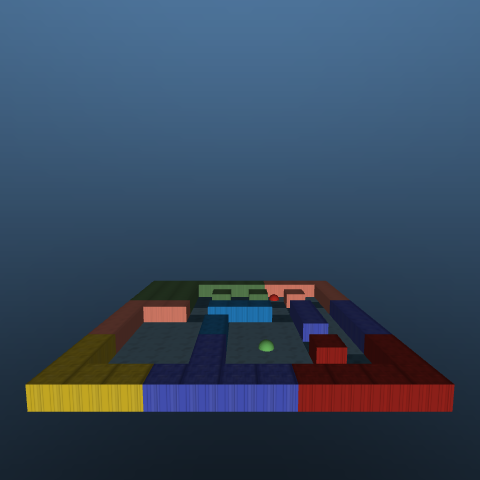}
        \caption{M9$\times$9-DET}
    \end{subfigure}
    \begin{subfigure}[b]{0.25\textwidth}
        \centering
        \includegraphics[width=\textwidth, trim=13 13 13 13, clip]{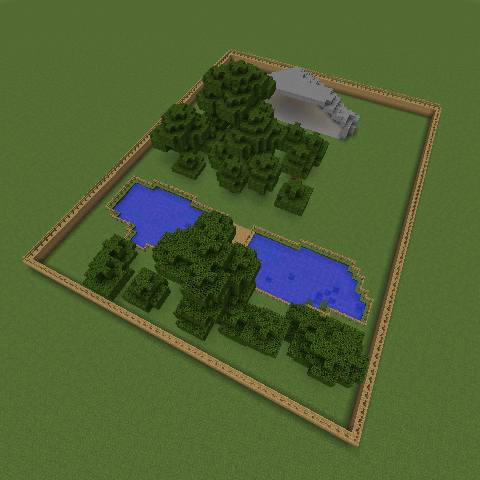}
        \caption{MC-DET}
    \end{subfigure}
    \caption{\small High-angle perspective views of the three evaluation environments. These renderings provide an intuitive, three-dimensional understanding of the maze layouts that complements the 2D top-down maps in the main text.}
    \label{fig:high_view}
\end{figure}

\section{Ablation Studies}

We conduct ablation studies to validate the contribution of our core components and design choices. 
Specifically, we examine four aspects: (1) the necessity of the two core loss terms, (2) the role of the projection head, (3) the impact of latent dimension, and (4) the effect of critical design choices on model performance. 
The detailed results are provided in the supplementary material.

\section{Additional Rollout Visualizations}\label{app:horizon}
\begin{figure}[h!]
    \centering
    \begin{subfigure}[t]{\textwidth}
        \centering
        \includegraphics[width=\linewidth]{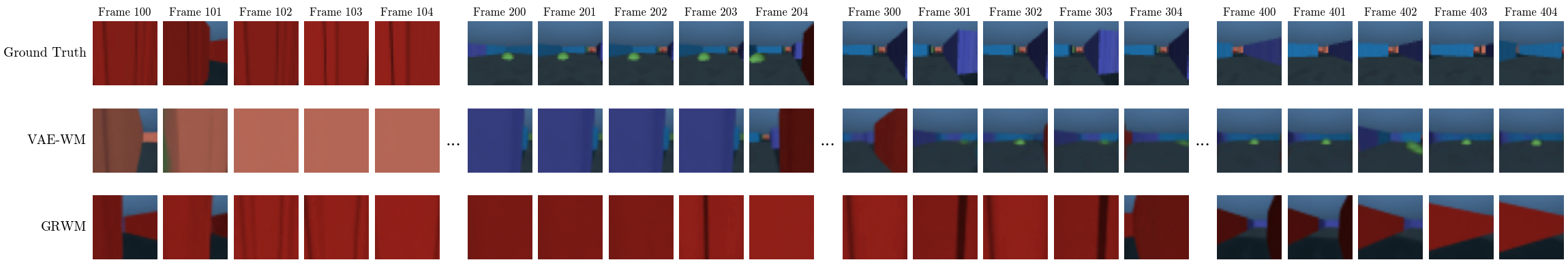}
        \includegraphics[width=\linewidth]{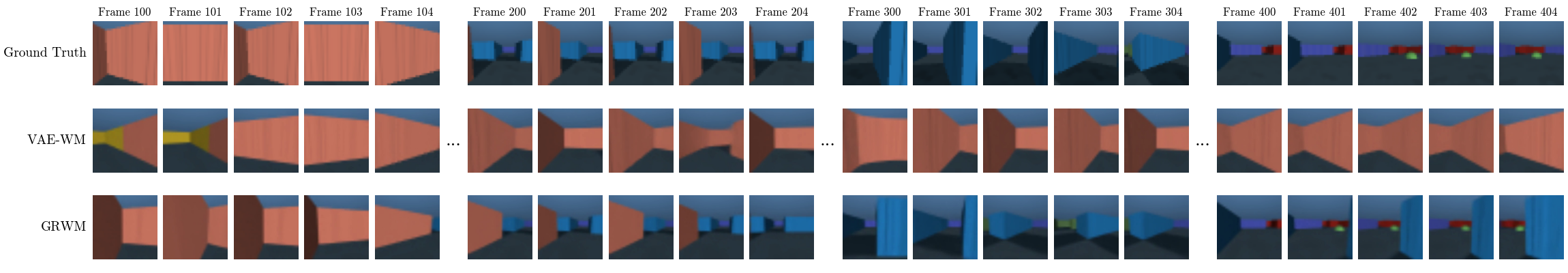}
        \includegraphics[width=\linewidth]{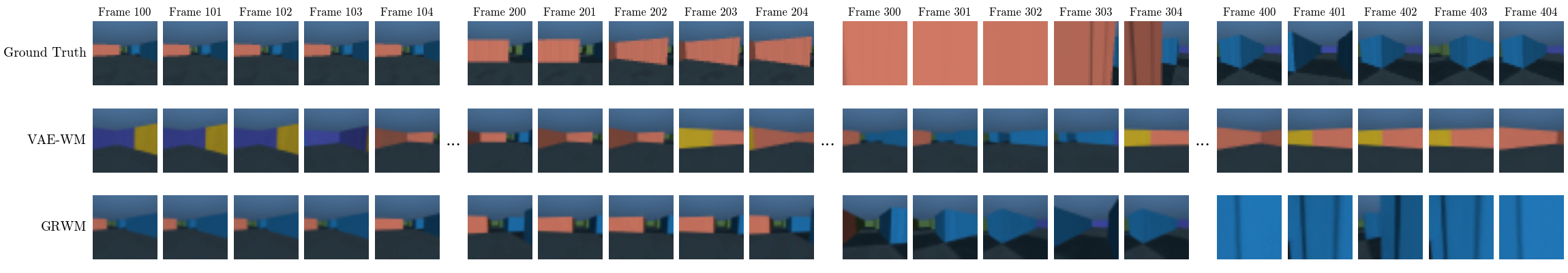}
        \includegraphics[width=\linewidth]{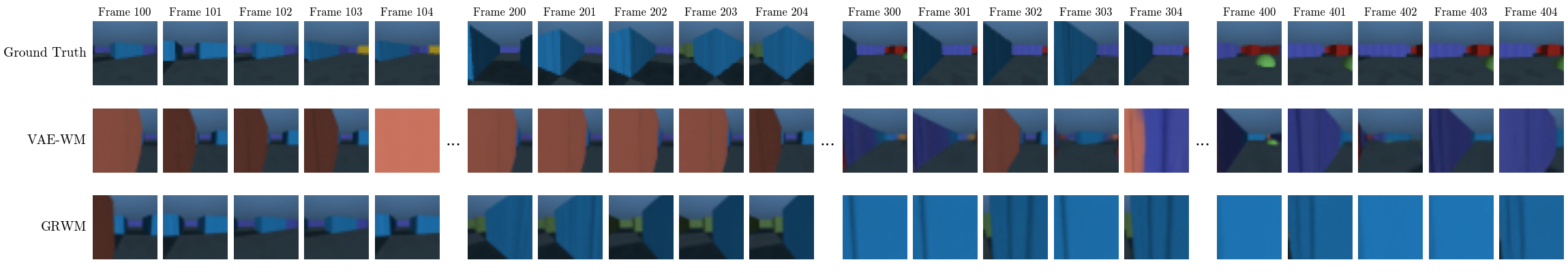}
        \includegraphics[width=\linewidth]{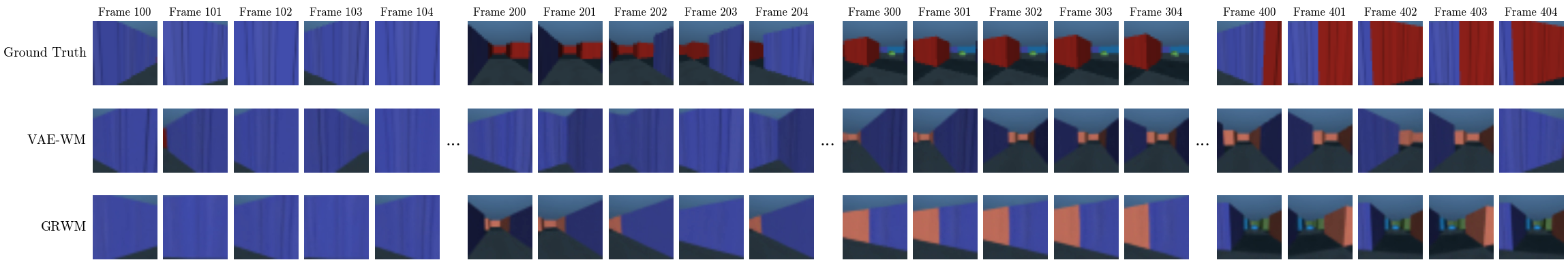}
        \includegraphics[width=\linewidth]{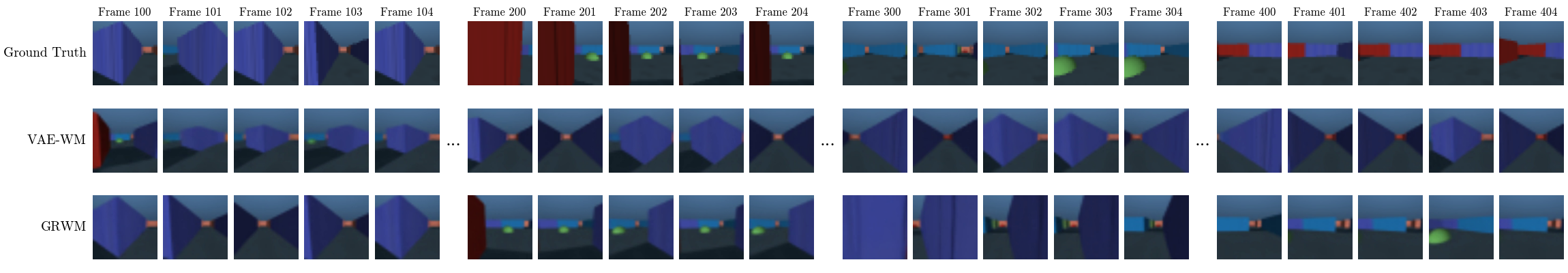}
        \includegraphics[width=\linewidth]{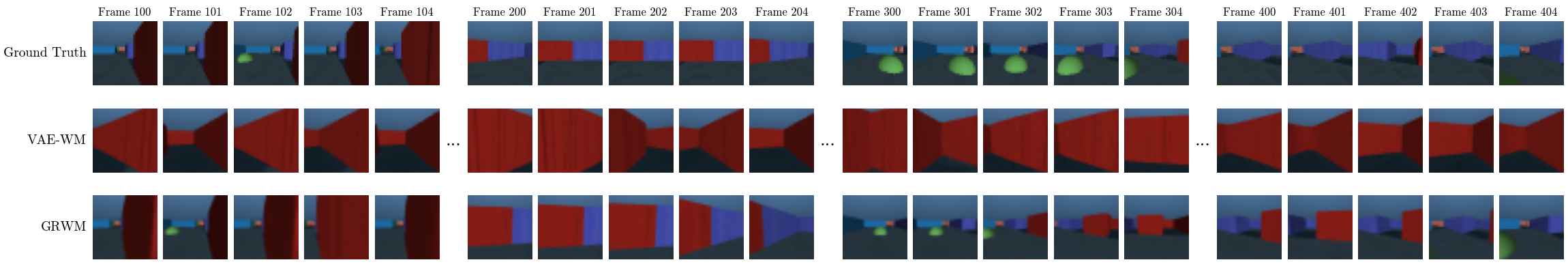}
    \end{subfigure}

    \caption{Visualization of generated frames at multiple time points from a single starting state. We show frames near steps 100, 200, 300, 400, sampled randomly — no cherry-picking. Our method significantly outperforms the VAE baseline: while the VAE predictions are already inaccurate at 100 steps, our model maintains high fidelity at this horizon. In some case, our method can occasionally produce accurate predictions even at 400 steps, demonstrating the improved consistency of the latent-space trajectories with the true environment.}
    \label{fig:long_rollouts_full}
\end{figure}

\begin{figure}[h!]
    \centering
    \begin{subfigure}[t]{\textwidth}
        \centering
        \includegraphics[width=\linewidth]{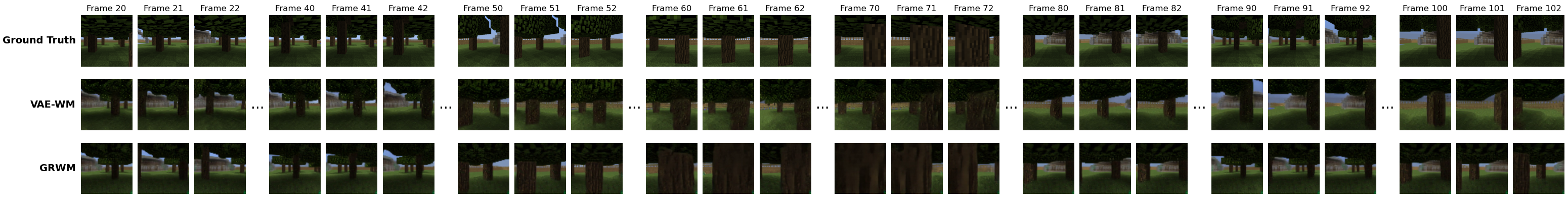}
        \includegraphics[width=\linewidth]{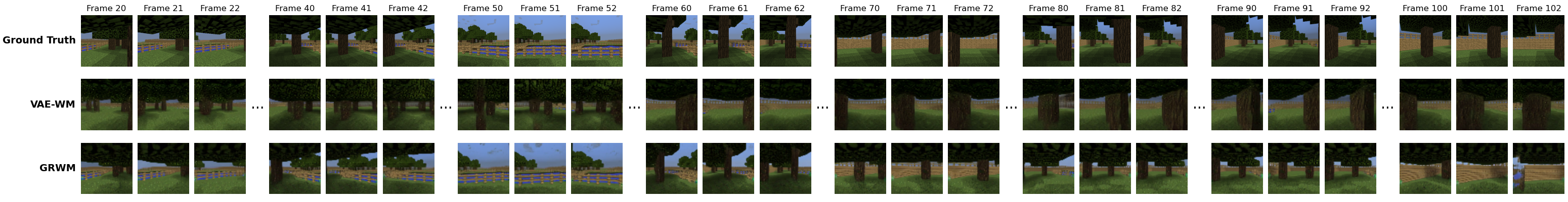}
        \includegraphics[width=\linewidth]{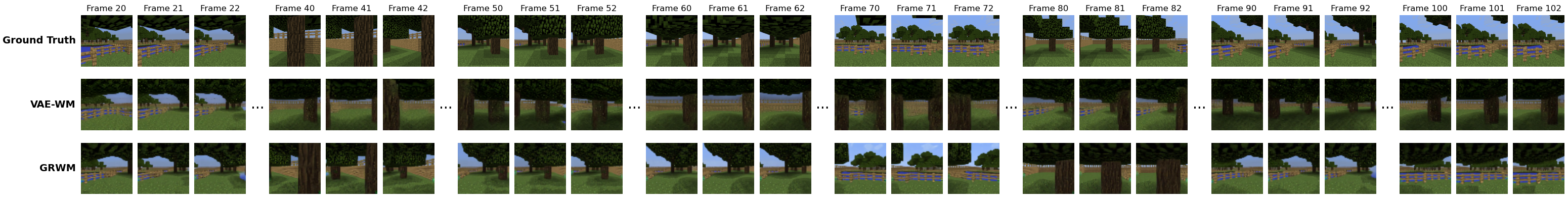}
        \includegraphics[width=\linewidth]{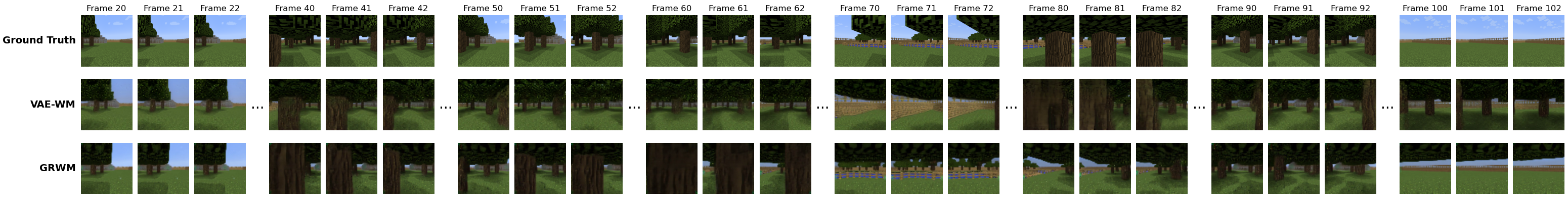}
        \includegraphics[width=\linewidth]{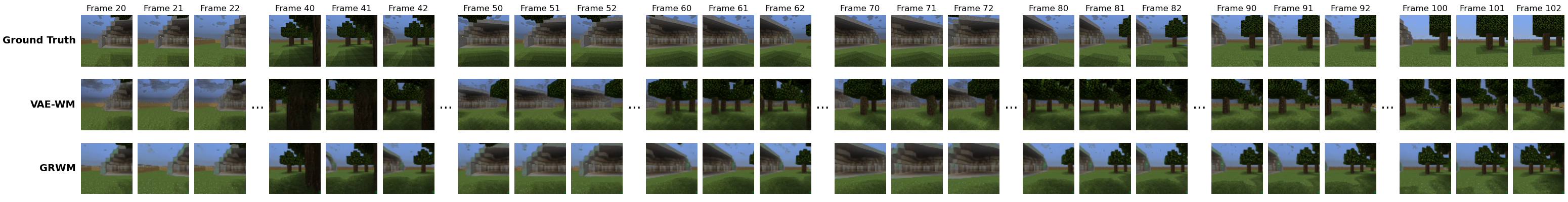}
        \includegraphics[width=\linewidth]{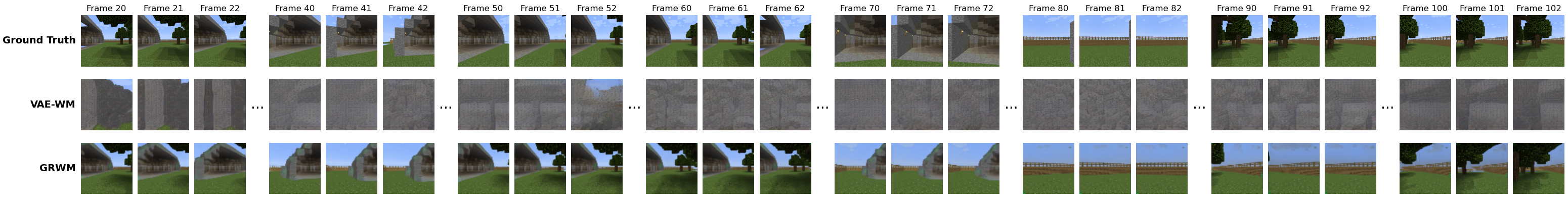}
        \includegraphics[width=\linewidth]{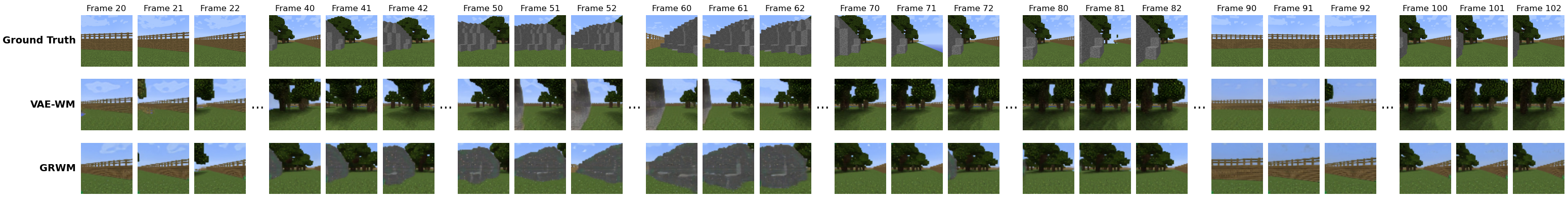}
        \includegraphics[width=\linewidth]{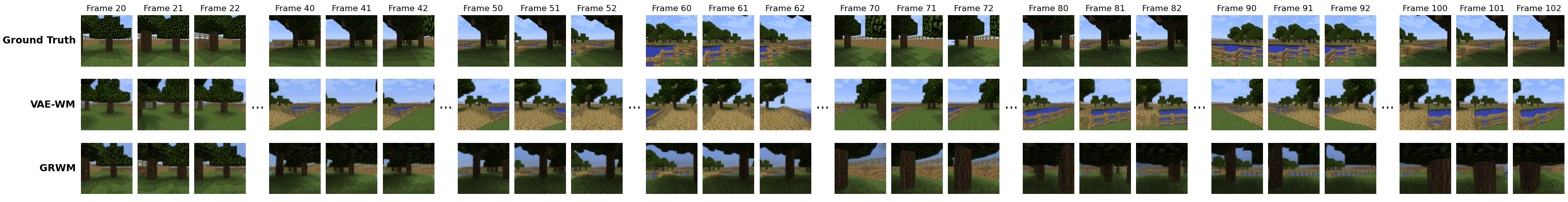}
        \includegraphics[width=\linewidth]{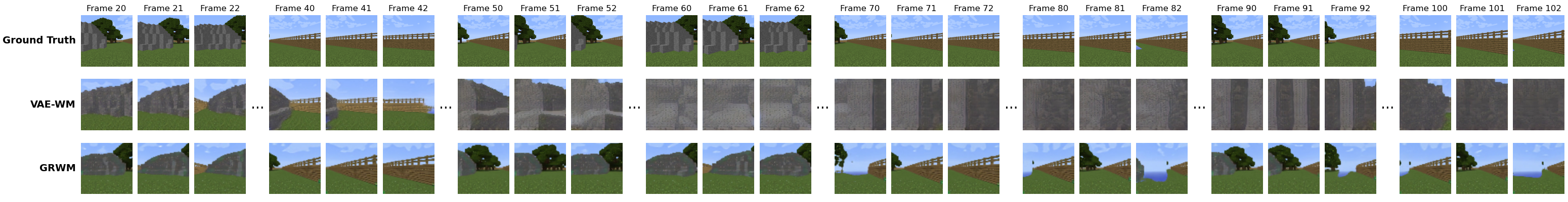}
        \includegraphics[width=\linewidth]{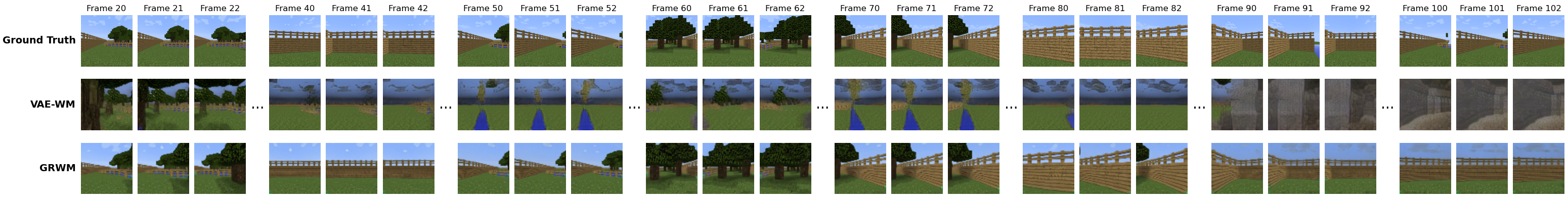}
    \end{subfigure}

    \caption{Visualization of generated frames from the MC-DET sequence.}
    \label{fig:MC_sequence}
\end{figure}

We provide additional rollout visualizations for both the M9x9-DET environment in Figure~\ref{fig:long_rollouts_full} and the MC-DET environment in Figure~\ref{fig:MC_sequence}. As shown in the M9x9-DET results (Figure~\ref{fig:long_rollouts_full}), our method significantly outperforms the VAE baseline: while the VAE predictions are already inaccurate at 100 steps, our model maintains high fidelity at this horizon. In some cases, our method can occasionally produce accurate predictions even at 400 steps, demonstrating the improved consistency of the latent-space trajectories with the true environment. The MC-DET results (Figure~\ref{fig:MC_sequence}) further confirm this robustness on complex trajectories. These results are not cherry-picked; the figure shows samples from randomly selected starting points, illustrating the typical performance of both methods over long-horizon rollouts.

\end{document}